\documentclass[a4paper,fleqn]{cas-dc}

\makeatletter
\newcommand{\Autoref}[1]{%
  \begingroup
  \def\sectionautorefname{Section}%
  \def\subsectionautorefname{Subsection}%
  \def\subsubsectionautorefname{Subsubsection}%
  \def\figureautorefname{Figure}%
  \def\tableautorefname{Table}%
  \def\equationautorefname{Equation}%
  \autoref{#1}%
  \endgroup
}
\makeatother

\usepackage[numbers]{natbib}
\usepackage{lipsum}
\usepackage{siunitx}
\usepackage{subcaption}
\usepackage{makecell}
\usepackage{multirow}
\usepackage{placeins}
\usepackage[switch]{lineno}

\definecolor{LightYellow}{RGB}{255,255,128}
\usepackage{soul}
\setul{-1.5ex}{2ex}
\setulcolor{LightYellow}

\usepackage{courier}
\usepackage{stfloats}
\usepackage[export]{adjustbox}
\usepackage{caption}

\usepackage{silence}
\def\tsc#1{\csdef{#1}{\textsc{\lowercase{#1}}\xspace}}
\tsc{WGM}
\tsc{QE}
\tsc{EP}
\tsc{PMS}
\tsc{BEC}
\tsc{DE}
\begin{document}
\let\WriteBookmarks\relax
\def\floatpagepagefraction{1}
\def\textpagefraction{.001}

\shorttitle{Multimodal Hand Gesture Recognition Using Quey-Based Transformers}    

\shortauthors{F. Del Pup et~al.}  

\title[mode = title]{Multimodal Surface EMG Hand Gesture Recognition Using Query-Based Transformers for Prosthetic Control}  



% ========== FIRST AUTHOR ==========
\author[1,2]{Federico {Del Pup}}[
    type=author,
    auid=000,
    bioid=1,
    orcid=0009-0004-0698-962X
]
\cormark[1]


\ead{federico.delpup@unipd.it}


\credit{
    Conceptualization,
    Methodology,
    Software,
    Formal Analysis,
    Writing - Original Draft
}

% ========== SECOND AUTHOR ==========
\author[2,3]{Elisa {Tentori}}[
    type=author,
    auid=001,
    bioid=2,
    orcid=0000-0002-4755-1577
]
\ead{elisa.tentori@unipd.it}
\credit{
    Formal Analysis,
    Software,
    Visualization,
    Investigation,
    Writing - review and editing
}

% ========== THIRD AUTHOR ==========
\author[2,4,5]{Manfredo {Atzori}}[
    type=author,
    auid=002,
    bioid=3,
    orcid=0000-0001-5397-2063
]
\ead{manfredo.atzori@unipd.it}
\credit{
    Supervision,
    Funding acquisition,
    Project administration,
    Writing - review and editing
}

% -------------------------
%    AFFILIATION SECTION
% -------------------------
\affiliation[1]{
    organization={Department of Information Engineering, University of Padua}, 
    city={Padua},
    postcode={35131}, 
    country={Italy}
}

\affiliation[2]{
    organization={Padova Neuroscience Center, University of Padua}, 
    city={Padua},
    postcode={35129}, 
    country={Italy}
}

\affiliation[3]{
    organization= {Department of  Biomedical Sciences, University of Padua},
    city={Padua},
    postcode={35131},
    country={Italy}
}

\affiliation[4]{
    organization={Department of Neuroscience, University of Padua}, 
    city={Padua},
    postcode={35121}, 
    country={Italy}
}

\affiliation[5]{
    organization={Information Systems Institute, University of Applied Sciences Western Switzerland (HES-SO Valais)}, 
    city={Sierre},
    postcode={3960}, 
    country={Switzerland}
}

\cortext[1]{Corresponding author. \\ This document is the results of the research project by the European Union’s Horizon Europe research and innovation programme under Grant agreement no 101137074 - HEREDITARY.}

\begin{abstract}
%Background:
Hand gesture recognition via surface electromyography (sEMG) is fundamental to human-machine interaction and prosthetic control.
In this field, deep learning approaches have become the gold standard.
However, current architectures struggle to scale; model performance typically decreases as the number of hand movements increases.
%
%Motivation:
Performance degradation is tied to the increased statistical complexity of decoding expanded gesture sets and compounded by the limitations of state-of-the-art approaches, which primarily rely on low-latency unimodal convolutional architectures.
Convolutions operate locally, limiting model's ability to capture long-range sequential patterns.
Unimodal setups cannot leverage complementary information from coordinated signals characterizing movement execution, such as inertial and eye-tracking data.
These limitations motivate architectures that integrate local and global features across multimodal physiological sequences.
%
%Objective:
To bridge this gap, this study introduces EMG-CrossFormer, an end-to-end hybrid convolutional-transformer for seamless multimodal integration.
EMG-CrossFormer combines representations from an arbitrary number of unimodal encoders through cascaded cross-attention fusion layers, and decodes the fused representations using learnable gesture queries.
%
%Methods:
EMG-CrossFormer was evaluated on four NinaPro datasets (DB2, DB3, DB7, and DB10) and benchmarked against six state-of-the-art models using increasing number of modalities.
%
%Results:
Using only sEMG signals, EMG-CrossFormer achieved mean accuracies of 72.33\%, 52.48\%, 79.16\%, and 73.49\% on DB2, DB3, DB7, and DB10, respectively, consistently achieving the highest performance.
Incorporating inertial signals improved performance to 90.66\%, 80.40\%, 92.79\%, and 92.06\%.
%
%Conclusion:
These results show that joint local-global feature modeling improves sEMG-only decoding and that multimodal fusion substantially amplifies this benefit, underscoring the value of both design principles for complex hand gesture recognition in prosthetics.
\end{abstract}

%\item  This work presents EMG-CrossFormer, a hybrid convolutional-transformer model.
%\item  EMG-CrossFormer can integrate an arbitrary number of input modalities.
%\item  EMG-CrossFormer was evaluated on four NinaPro datasets for hand gesture decoding.
%\item  EMG-CrossFormer improves both unimodal and multimodal decoding accuracy.
%\item  EMG-CrossFormer maintains reduced FLOPs and inference latency.

%\nocite{*}

% Keywords

\maketitle

% Main text
\section{Introduction}
\label{sec:introduction}

% Intro to HGR and importance for amputees
Surface electromyography (sEMG)-controlled upper-limb prostheses have been used clinically since the 1960s \cite{upperlimbrev}, and decades of subsequent research have progressively refined their control algorithms and improved their reliability.
These myoelectric devices have been shown to enhance the quality of life for trans-radial amputees by partially restoring limb motor functions \cite{semgissue2, semgissue3, semgissue1}.

% Commercial solutions
To date, commercial myoelectric solutions integrate pattern-recognition algorithms for automatic sEMG gesture recognition, typically supporting a limited set of functional grasps with high accuracy and minimal latency \cite{semgdevices}.
Concurrently, advances in the computational capabilities of embedded systems have driven a shift in the research community from classical machine learning approaches to deep learning paradigms \cite{DLTrio}.
However, this transition has not yet been reflected in commercial pattern-recognition systems, which still predominantly rely on classical algorithms.

% deep learning models underperform with high number of gesture
In research settings, several studies have reported outstanding decoding performance, frequently exceeding 90\% accuracy on commonly used gesture sets comprising up to 17 hand movements representative of activities of daily living \cite{mkcnn, emghandnet, lstm1, trahgr}.
However, current literature in this domain often suffers from a lack of reproducibility, complicating the direct benchmarking of different architectures under identical experimental settings.
Furthermore, common evaluation protocols frequently omit various movements identified as highly similar by quantitative taxonomies \cite{taxonomy},
simplifying the classification task, potentially inflating reported performance, and decreasing the usefulness of benchmark datasets.
These evaluation gaps make it difficult to critically assess the true benefits of deep learning frameworks over classical machine learning models, such as Random Forests (RF) or Support Vector Machines (SVM), which continue to offer highly competitive performance \cite{ninapro}.

% examples on low
%This limitation becomes particularly evident when deep learning models are scaled to handle a larger set of hand gestures.
These limitations of current deep learning approaches become particularly evident when models are scaled to larger sets of hand gesture.
Under these demanding scenarios, the performance of deep learning models drops significantly, often falling below that of machine learning classifiers.
For example, in \cite{shallowcnn}, the authors evaluated a shallow Convolutional Neural Network (CNN) against a set of 51 distinct hand movements from the Non Invasive Adaptive Prosthetics (NinaPro) dataset \cite{ninaprodb}, revealing a performance degradation of more than 10\% compared to a RF across both intact and amputee cohorts.

% Problems in common architectures
The reported reduction in performance highlights some limitations of the existing deep learning frameworks.
State-of-the-art networks rely predominantly on unimodal convolutional architectures optimized for low latency.
While conventional convolutions enable highly parallelized and efficient computations that keep system latency below the recommended real-time threshold of 125 ms \cite{latency}, they are inherently limited to extracting local features.
This local receptive field limits the model's ability to distinguish complex neural activation patterns from noise, thereby reducing decoding accuracy and preventing the model from fully exploiting the performance gains offered by deep neural networks.
In contrast, attention layers offer a complementary solution that allows sEMG deep learning models to capture longer temporal dynamics \cite{transformer}.
However, their integration into low-latency, resource-constrained models remains scarcely investigated.

Similarly, unimodal (sEMG-only) approaches prevent these architectures from effectively leveraging complementary information from other signal modalities, such as accelerometer data, which can help characterize both motor intent and the physical execution of movement.
Previous work has demonstrated that combining sEMG signals with accelerometer data can improve classification accuracy in amputee subjects to levels comparable to intact individuals \cite{semgacc, semgacc2}.
Consistent with these findings, the authors in \cite{nkdff} achieved comparable results with a novel multimodal convolutional model that integrates sEMG and accelerometer data within a compact, end-to-end framework.

% need to advance 
Despite the benefits highlighted, multimodal solutions integrating further modalities (e.g., eye tracking, scene videos) remain less investigated.
These limitations motivate the development of novel architectures capable of simultaneously extracting and combining local and global features across heterogeneous multimodal physiological sequences.

% contribution
\textit{\textbf{Contributions:}} To bridge this gap, this study introduces EMG-CrossFormer, an end-to-end, hybrid convolutional-transformer framework designed for seamless multimodal signal integration.
EMG-CrossFormer combines representations from an arbitrary number of unimodal encoders through cascaded cross-attention fusion layers.
Then, it decodes the fused representations via cross-attention with learnable gesture queries, inspired by consolidated query-based decoding approaches applied in other domains (e.g., DETR \cite{detr}).
%This structure enables scalable and robust gesture recognition.
EMG-CrossFormer was evaluated on four NinaPro databases (DB2, DB3, DB7, and DB10) and benchmarked against six state-of-the-art models using an incremental number of input modalities.
While this study focuses on the integration of sEMG with accelerometer and eye-tracking data, EMG-CrossFormer has been designed to be easily extended to additional data modalities, facilitating the development of multimodal architectures incorporating, for instance, scene camera or EEG data.

% paper structure
\textit{\textbf{Paper structure:}} the outline of this paper is as follows.
\Autoref{sec:methods} describes in detail the model architecture and the experimental setting.
\Autoref{sec:results} presents the comparative analysis between EMG-CrossFormer and the other selected models using an incremental number of data modalities.
\Autoref{sec:discussion} critically discusses the results, highlighting potential limitations and future directions.
Finally, a conclusion is drawn in \autoref{sec:conclusion}.

\section{Methods}
\label{sec:methods}

This section outlines the methodological design of the study, providing thorough details on the procedure and the architecture, and fostering the reproducibility of the results.
Specifically, it covers dataset selection, data preprocessing, model architecture, training hyperparameters, performance evaluation, and statistical analysis.

\subsection{Datasets}
The proposed EMG-CrossFormer model was evaluated using four distinct datasets from the public NinaPro repository\footnote{[Online] Available: \url{https://ninapro.hevs.ch/}} \cite{ninapro}: datasets 2, 3, 7, and 10, referred to as DB2, DB3, DB7, and DB10, respectively.
\autoref{tab:ninapro_datasets} summarizes their main acquisition configurations.

The datasets were selected to ensure a comprehensive evaluation of decoding performance across heterogeneous cohorts of both intact and trans-radial amputee subjects, while also investigating increasing levels of sensor multimodality.
%\autoref{tab:ninapro_datasets} summarizes their main acquisition configurations.
\noindent DB2, DB3, and DB7 provide synchronized surface electromyography (sEMG) and tri-axial accelerometer recordings acquired with a 12-channel Delsys Trigno\texttrademark \ system (Delsys, Natick, MA, USA).
They include a broad set of hand-motion and grasping movements collected from both intact subjects (DB2, DB7) and trans-radial amputee subjects (DB3, DB7).
Movements were organized into three sessions, referred to as Exercises B, C, and D, comprising 17, 23, and 9 movements, respectively.
%The experimental protocol consisted of 6 repetitions per movement, 
Each gesture was repeated six times; active trials lasted $5$ seconds and were followed by $3$ seconds of rest.
DB7 only includes Exercises B and C. Therefore, Exercise D was excluded from DB2 and DB3 to harmonize the label space across the three datasets, yielding 40 discrete gestures plus the resting state. 
sEMG signals were acquired at a sampling rate of $2$~kHz, whereas accelerometer signals were sampled at $148$~Hz and upsampled by the original authors to match the sEMG sampling rate.
All subjects were included in the analysis except for two DB3 subjects, for whom the number of electrodes was reduced due to insufficient residual limb space, according to the NinaPro authors’ usage notes~\cite{ninapro}.

\noindent DB10 includes $10$ grasping movements and the resting state, selected from exercises B and C based on activities of daily living. Data were collected from a cohort of $30$ intact subjects and $15$ trans-radial amputee subjects \cite{db10}.
Each movement repetition lasted between $5$ and $6$ seconds and was followed by $4$ seconds of rest.
DB10 was acquired with three recording modalities at the same time: sEMG, accelerometer and gaze signals.
sEMG signals were recorded using a 12-channel Delsys Trigno\texttrademark \ Wireless system at a sampling frequency of $1926$~Hz.
Upper-limb kinematics were captured via tri-axial accelerometers at a native sampling rate of $148$~Hz, and eye-gaze dynamics were recorded using Tobii Pro Glasses 2 at $100$~Hz.
Both accelerometer and gaze signals were upsampled by the original authors to match the sEMG sampling rate and ensure temporal alignment across modalities~\cite{db10}.
Following the authors’ usage notes, six subjects were discarded due to poor signal quality, and repetitions marked as unreliable by the original authors were excluded~\cite{db10}.

\begin{table*}[t]
\centering
\small
\caption{Main acquisition configurations of the NinaPro datasets used in this work.}
\setlength{\tabcolsep}{3pt}
\renewcommand{\arraystretch}{1.1}
\begin{tabular}{llllllll}
\toprule
\makecell[l]{\textbf{Dataset}\\\textbf{Name}} &
\makecell[l]{\textbf{Subjects}$^*$} &
\textbf{Device} &
\textbf{Channels} &
\makecell[l]{\textbf{sEMG}\\\textbf{sampling}\\\textbf{rate [Hz]}} &
\makecell[l]{\textbf{Modalities used}} &
\makecell[l]{\textbf{Movements}} &
\makecell[l]{\textbf{Repetitions}}\\
\midrule
DB2  & 40 & Delsys Trigno & 12 & 2000 & sEMG + ACC & 40 + rest & 6 \\
DB3$^{**}$  & 11(A) & Delsys Trigno & 12 & 2000 & sEMG + ACC & 40 + rest & 6 \\
DB7  & 20 + 2(A) & \makecell[l]{Delsys Trigno (Wireless)} & 12 & 2000 & sEMG + ACC & 40 + rest & 6 \\
DB10$^{***}$ & 30 + 15(A) & \makecell[l]{Delsys Trigno (Wireless)} & 12 & 1926 & \makecell[l]{sEMG + ACC + Gaze} & 10 + rest & 10 \\
\bottomrule
\multicolumn{8}{l}{$^{*}$\footnotesize A stands for Amputees; \small  $^{**}$\footnotesize Subjects 6 and 7 were discarded according to usage notes provided in \cite{ninapro}}\\
[-0.25em]
\multicolumn{8}{l}{$^{***}$\footnotesize Six subjects and unreliable repetitions were discarded according to usage notes provided in \cite{db10}}
\end{tabular}
\label{tab:ninapro_datasets}
\end{table*}

%------------------------------------ Preprocessing ---------------------------------------%

\subsection{Data Preprocessing}

Raw sEMG signals were preprocessed using the following pipeline:

\WarningsOff
\begin{enumerate}[$\bullet$]
    \item \textit{scale conversion}: sEMG signals were converted to millivolts to improve numerical stability and avoid representation issues during mixed-precision (FP16) training.
    \item \textit{DC component removal}: the channel mean was subtracted from each sEMG channel, which is equivalent to removing the DC component.
    \item \textit{Filtering}: sEMG signals were filtered using a second-order (12 dB/oct) forward-backward Butterworth band-pass filter with cutoff frequencies of $20$~Hz and $500$~Hz. The filter order and cutoff frequencies were selected according to guidelines for biomechanical and clinical applications \cite{semgfilt}. Power-line noise and its harmonics had already been removed by the original authors using a Hampel filter \cite{Hampel}.
    \item \textit{Resampling}: sEMG signals were resampled to $1$~kHz to reduce memory footprint and the number of floating-point operations (FLOPs) required by the deep learning architectures.
    \item \textit{Window extraction}: sEMG data were partitioned into windows of $100$, $150$, or $200$~ms with a $10$\% shift (corresponding to $90$\% overlap) to increase the number of samples.
    \item \textit{Rest class balancing}: windows belonging to the resting class were undersampled to match the gesture class ratios.
\end{enumerate}
\WarningsOn

Accelerometer data underwent only resampling and window extraction to preserve temporal consistency.
Gaze signals, represented as $(x,y)$ coordinates in image space, were processed in the same way after short intervals of missing values were imputed by linear interpolation, following the guidelines provided in \cite{gazeinterp}.

%---------------------------------------------------------------------------%

\subsection{EMG-CrossFormer Architecture}

\begin{figure*}[t]
\centering
\includegraphics[trim={1cm 2cm 1cm 2cm}, clip, width=\linewidth]{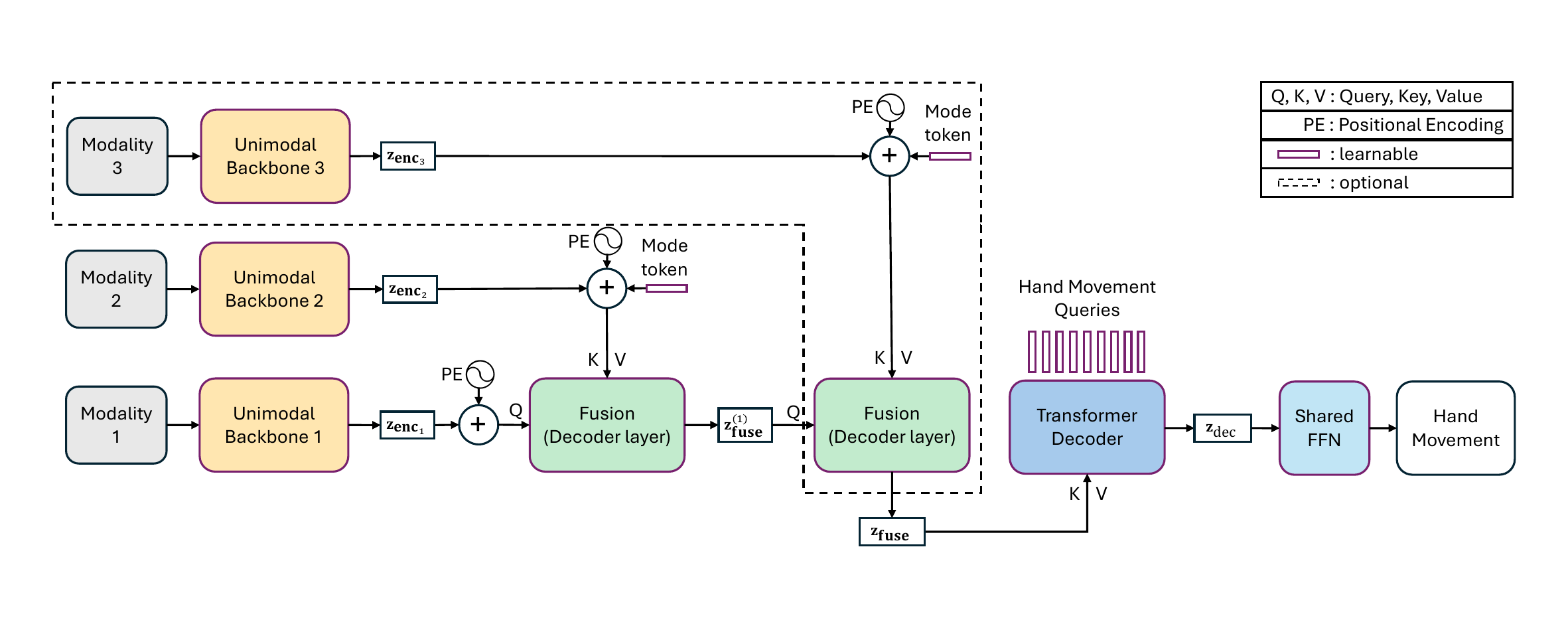}
\caption{
Schematic representation of EMG-CrossFormer. The model combines representations ($\mathbf{z}_{\text{enc}_m}$) from unimodal backbones through cascaded fusion layers (transformer decoder layers). The transformer decoder block decodes the fused representation ($\mathbf{z}_{\text{fuse}}$) using learnable hand movement queries. A final shared FFN outputs the hand movement predictions from the decoded representation ($\mathbf{z}_{\text{dec}}$).
EMG-CrossFormer can integrate an arbitrary number of modalities.
To illustrate this extensibility, the third modality branch and its corresponding fusion layer are shown as optional (dashed box).
}
\label{fig:emgcrossformer}
\end{figure*}

EMG-CrossFormer is a hybrid convolutional-transformer model designed to jointly process and fuse multiple input modalities (signals in this study) for hand-movement decoding.
As illustrated in \autoref{fig:emgcrossformer}, the architecture consists of four main modules:
modality-specific backbones, which map each input modality to a compact sequence of feature tokens; 
a fusion module that combines the token sequences through a cascade of cross-attention fusion blocks;
a transformer decoder with learnable gesture queries;
and a two-layer feed-forward network (FFN) for hand-gesture recognition.

The model is highly flexible and was designed to be easily adaptable to different experimental setups, including different types and numbers of input modalities.
Unimodal backbones may differ across modalities, provided that they output representations in the form of a sequence of tokens.
Furthermore, the number of modalities can also vary by adding or removing fusion blocks in the cascade.

A PyTorch \cite{pytorch} implementation of the model is provided in the openly available source code\footnote{[Online] Available: https://github.com/deepPNClab/emg-crossformer}.
Initialization hyperparameters are also reported in the Supplementary Materials.

%---------------------------------------------------------------------------%

\subsubsection{Unimodal Backbone}

Let $m$ denote the input modality, $C_m$ the number of channels and $W_m$ the number of samples within the input window. Given an input multi-channel signal $\mathbf{x}_{\text{sig}_m}~\in~\mathbb{R}^{C_m \times W_m}$, the $m$-th unimodal backbone generates a compact representation $\mathbf{z}_{\text{enc}_m} \in \mathbb{R}^{L_m \times d}$ (\autoref{fig:emgcrossformer}),  where $L_m$ is the sequence length (number of output tokens) and $d$ is the token embedding dimension. $L_m$ may vary across modalities, particularly when the corresponding signals have different sampling rates. By contrast, $d$ is kept fixed across modalities to enable the cross-attention in the fusion and decoding modules. 

To assess the architecture's adaptability to different encoder designs, two unimodal backbone implementations were evaluated: a 1D depthwise convolutional encoder and a 2D multi-scale convolutional encoder.

The 1D depthwise implementation, based on \cite{transformeeg}, extracts channel-specific temporal features through stacked depthwise convolutions and average pooling, with low parameter cost.
Each block doubles the feature dimension while halving the temporal resolution.
Because convolutions are depthwise, this implementation maintains a low parameter count and restricts feature extraction to temporal patterns, leaving the modeling of inter-channel patterns to subsequent stages of the architecture.

The 2D multi-scale implementation, based on \cite{mkcnn}, treats the multi-channel signals as a single-channel pseudo-image and applies parallel convolutional branches with different kernel sizes, to capture patterns at multiple frequency scales.
The extracted features are subsequently combined through separable convolutions to generate a compact representation while reducing the parameter count.
Compared to the 1D depthwise backbone, the 2D multi-scale architecture extracts spatio-temporal features at the cost of increased parameters and computational load.

To integrate these backbones into the EMG-CrossFormer architecture while preserving the architectural designs proposed by the original authors, a learnable linear projection layer is appended to each encoder to map its output into a shared $d$-dimensional embedding space.
The resulting EMG-CrossFormer variants are denoted as $\text{EMG-CF}_{\text{1D}}$ and $\text{EMG-CF}_{\text{2D}}$, respectively.

%---------------------------------------------------------------------------%
\subsubsection{Fusion Layers}
\label{subsec:fusion}

%To seamlessly combine an arbitrary number of input modalities $M$, t
The fusion stage combines a variable number $M$ of input modalities through a sequential cascade of non-causal transformer decoder layers.
Let $\mathbf{z}_{\text{enc}_m}~\!\in~\!\mathbb{R}^{L_m \times d}$ represent the compact tokenized sequence generated by the $m$-th unimodal backbone, with $m~\!\in~\!\{1, \dots, M\}$. Here, $m\!=\!1$ denotes the primary sEMG modality, while $m\!>\!1$ denotes subsequent auxiliary signals.

Before fusion, a sinusoidal positional encoding $E_{\text{pos}}^m \in \mathbb{R}^{L_m \times d}$ is injected into each unimodal feature set.
For auxiliary modalities, a learnable modality-specific bias, defined as the mode token $\mathbf{t}_{\text{mode}}^{m} \in \mathbb{R}^{1 \times d}$, is also added directly to the sequence (see "Mode token" in \autoref{fig:emgcrossformer}).
This token acts as an indicator to help fusion blocks contextualize non-sEMG signal representations.
For each modality, the processed representation $\bar{\mathbf{z}}_{\text{enc}_m}$ is defined as:

\begin{equation}
\bar{\mathbf{z}}_{\text{enc}_m} = \begin{cases} 
\mathbf{z}_{\text{enc}_m} + E^m_{\text{pos}} & \text{if } m = 1 \\ 
\mathbf{z}_{\text{enc}_m} + E^m_{\text{pos}} + \mathbf{1}_{L_m}\mathbf{t}^m_{\text{mode}} & \text{if } m > 1
\end{cases}
\end{equation}

\noindent where $\mathbf{1}_{L_m} = (1,\dots,1)^T \in \mathbb{R}^{L_m \times 1}$.

The mixing of representations across modalities is performed via a directional cascade, as schematized in \autoref{fig:emgcrossformer}.
For notational reasons, the representations of the first primary modality are considered the initial state of the fusion stage, such that $\mathbf{z}_{\text{fuse}}^{(1)}~\!=~\!\bar{\mathbf{z}}_{\text{enc}_1}$.
For each subsequent auxiliary modality $m = 2, \dots, M$, the $(m-1)$-th fusion block applies the set of operations of a non-causal transformer decoder layer.\\
More formally, the output of the $m$-th block in the cascade is described as:

\begin{equation}
    \mathbf{z}_{\text{fuse}}^{(m)} = \text{LN}\Big( \hat{\mathbf{z}}^{(m)} + \text{FFN}\big(\hat{\mathbf{z}}^{(m)}\big) \Big),
\label{eq:decoder}
\end{equation}
where
\begin{equation}
\begin{aligned}
    \hat{\mathbf{z}}^{(m)} &= \text{LN}\Big( \tilde{\mathbf{z}}^{(m)} + \text{MHA}\big(\tilde{\mathbf{z}}^{(m)}, \bar{\mathbf{z}}_{\text{enc}_m}, \bar{\mathbf{z}}_{\text{enc}_m}\big) \Big) \\
    \tilde{\mathbf{z}}^{(m)} &= \text{LN}\Big( \mathbf{z}_{\text{fuse}}^{(m-1)} + \text{MHA}\big(\mathbf{z}_{\text{fuse}}^{(m-1)}, \mathbf{z}_{\text{fuse}}^{(m-1)}, \mathbf{z}_{\text{fuse}}^{(m-1)}\big) \Big)
\end{aligned}
\end{equation}

Here, $\text{LN}$ denotes the layer normalization, $\text{FFN}$ a pointwise feed-forward network, and $\text{MHA}$ the multi-head attention mechanism.
For an in-depth description of the multi-head attention mechanism, the reader is referred to the original implementation of the transformer model \cite{attention}.

Following the final cascade layer, the module outputs a highly descriptive, multimodal representation of the input signals $\mathbf{z}_{\text{fuse}}~\!=~\!\mathbf{z}_{\text{fuse}}^{(M)}~\!\in~\!\mathbb{R}^{L_1 \times d}$.
This representation fully compresses the dynamics of all available modalities while preserving the native token sequence length $L_1$ of the underlying sEMG driver.

In this study, transformer decoder layers were initialized with an embedding dimension $d=128$, 8 heads, SwiGLU activation \cite{swiglu}, and no dropout.

\subsubsection{Query-based decoder}
The fused multimodal representation $\mathbf{z}_{\text{fuse}} \in \mathbb{R}^{L_1 \times d}$ is fed into a query-based transformer decoder.
Following the formulation originally introduced for object detection tasks in computer vision \cite{detr}, this module implements the standard transformer decoder operations described in \autoref{eq:decoder}, using a set of learnable query embeddings as decoder inputs.
Analogously, EMG-CrossFormer uses learnable hand gesture queries, each encoding a candidate hand movement.
Through self-attention among queries and cross-attention over $\mathbf{z}_{\text{fuse}}$, the model jointly reasons over all gestures, capturing pairwise relationships while using the multimodal fused representation as contextual information.
In this study, the number of layers in the query-based decoder was set to 4, with each layer initialized using the same hyperparameters adopted for the fusion stage layers.

\subsubsection{Feed-forward network}
Final predictions are produced by a shared two-layer Feed-Forward Network (FFN) applied independently to each decoded hand gesture query.
The hidden layer has dimensionality $d=128$ and is followed by a ReLU activation.
The FFN outputs class logits for each query, which are subsequently normalized with a softmax function to obtain class probabilities.

\subsubsection{Model configuration}
In the unimodal setting, only sEMG data are used. To apply the fusion architecture within this setting, the sEMG signals are divided into two model input modalities: forearm channels, corresponding to the first eight channels, and upper-arm channels, corresponding to the last four channels.
The forearm channels are provided as the primary input modality, while the upper-arm channels are provided as an auxiliary input modality, potentially allowing the model to learn muscle synergies \cite{synergy, synergy2}.
In multimodal settings, the model receives sEMG together with accelerometer signals and, when available, gaze information. In this case, sEMG is kept as a single primary input modality, rather than being split into forearm and upper-arm channels. Accelerometer signals define the second input modality, while gaze information is used as the third input modality when present.
Additional details regarding model implementation, auxiliary training outputs, and hyperparameter selection are provided in the Supplementary Materials.

\subsection{Implementation details}
Deep learning models were implemented and trained using PyTorch \cite{pytorch}, while conventional machine learning models were implemented with Scikit-learn \cite{scikit-learn}. 
Statistical analyses were performed using SciPy \cite{scipy} and statsmodels \cite{statsmodels}.
Figures were generated with Seaborn \cite{seaborn}.
Experiments were conducted on the Department of Neuroscience computing cluster equipped with four NVIDIA A30 GPUs.
Further implementation details are available in the open-source codebase.

\subsubsection{Data partition}
Results were obtained using an intra-subject evaluation protocol, consistent with the domain \cite{nkdff, trahgr}, where prosthetic devices are required to work robustly on a specific subject.
In this setting, models are trained and evaluated on different repetitions of hand movements from the same subject.
Specifically, repetitions 1, 3, 4, and 6 were assigned to the training set, while repetitions 2 and 5 were assigned to the test set.
For DB10, additional repetitions (7 and 8) were included in the training set.

\subsubsection{Model comparison}
\label{subsec:modelcomp}

To provide a fair benchmark, four representative deep learning models were selected for comparison: Shallow CNN \cite{shallowcnn}, ResNet-1D \cite{resnet1d}, Multi-Scale Convolutional Neural Network (MKCNN) \cite{mkcnn}, and Narrow Kernel Dual-view Feature Fusion Convolutional Neural Network (NKDFF) \cite{nkdff}.
These models were selected because they achieve competitive performance on sEMG-based hand gesture recognition while representing different architectural paradigms, ranging from lightweight convolutional networks to residual and multi-scale architectures.
Moreover, their implementation details are sufficiently documented to enable faithful reproduction. NKDFF also provides a multimodal variant that incorporates accelerometer data.
Although additional recent models were considered, they were not included because insufficient implementation details, the absence of open-source code, or substantial differences in the original experimental protocols prevented a reliable reproduction and fair comparison.

The open-source codebase was designed to be extensible, enabling the straightforward integration of additional models and future benchmark extensions.
This design choice is intended to encourage community contributions to the source code, support the expansion of the benchmark analysis, and promote fair and reproducible evaluation of novel approaches.

Traditional machine learning models were also included in the evaluation, as they remain competitive for sEMG-based hand gesture recognition \cite{shallowcnn}, as discussed in \Autoref{sec:introduction} and further confirmed in \Autoref{sec:results}.
Specifically, a RF classifier and an SVM were evaluated.
%which according to literature provide consistently high performance in the domain \cite{shallowcnn}. %redundant with 219
These models were trained using the same set of handcrafted features listed in \cite{ninapro}.
Additional details on the features and the hyperparameter search grids are provided in the Supplementary Materials.

\subsubsection{Data Augmentation}
Data augmentation was incorporated during training to mitigate overfitting.
Specifically, a signal warping strategy was implemented to randomly stretch or compress portions of the input signal along the temporal axis, similarly to \cite{transformeeg, warp1}.
The procedure is defined as follows:
\begin{enumerate}
    \item The input window is divided into multiple segments.
    \item Up to half of these segments are randomly stretched, while the remaining segments are squeezed.
    \item A non-uniform temporal grid is constructed according to the selected stretch and compression operations. The grid values are determined by the stretch and compression strength hyperparameters.
    \item The input window is interpolated onto the non-uniform grid using the Piecewise Cubic Hermite Interpolating Polynomial (PCHIP) \cite{pchip}.
    \item The resulting signal is treated as uniformly sampled and resampled to the original temporal grid using PCHIP.
\end{enumerate}
Mini-batches were augmented with a probability of 70\% during training.
When signal warping was applied, the number of segments (4, 6, or 8), compression strength ($1.25$, $1.5$, or $2.0$), and stretch strength ($1.0$, $1.5$, or $2.0$) were randomly sampled from a predefined hyperparameter grid.
To improve computational efficiency, the same augmentation parameters were applied to all samples within a mini-batch through broadcasting along the batch dimension as well as to all input signals to preserve temporal consistency across modalities.

\subsubsection{Training loss}
\label{methods:loss}
EMG-CrossFormer was trained using a composite loss function designed to provide deep supervision across the encoding, fusion, and decoding stages. The loss jointly optimizes classification performance and embedding-space structure.
More formally, let $M$ denote the number of input modalities, and let:

\WarningsOff
\begin{enumerate}[$\bullet$] 
    \item $y$ be the ground-truth gesture label;
    \item $\mathbf{s}$ be the final FFN output logits;
    \item $\mathbf{s}_{\text{enc}_m}$ and $\mathbf{s}_{\text{fuse}}$ be auxiliary logits obtained from a linear projection of the global average pooling output of $\mathbf{z}_{\text{enc}_m}$ and $\mathbf{z}_{\text{fuse}}$, with $m \in {1,\dots,M}$;
    \item $\mathbf{z}_{\text{dec}}$ be the query-based decoder output;
    \item $\mathbf{h}$ be the ground-truth handcrafted sEMG features used by machine learning models (see Supplementary Materials);
    \item $\hat{\mathbf{h}}$ be the handcrafted sEMG feature estimates, obtained by applying a linear predictor of the global average pooling output of $\mathbf{z}_{\text{dec}}$.
\end{enumerate}
\WarningsOn

The applied training loss is defined as:
\begin{equation}
\begin{aligned}
\mathcal{L}_{total} &=
\lambda_{ce}\mathcal{L}_{CE}(\mathbf{s}, y) \\
& +
\lambda_{supcon}
\left[
\mathcal{L}_{SupCon}(\mathbf{z}_{\text{dec}}, y)
+
\sum_{i=1}^{M}
\mathcal{L}_{SupCon}(\mathbf{z}_{\text{enc}_i}, y)
\right] \\
& +
\lambda_{aux}
\left[
\mathcal{L}_{CE}(\mathbf{s}_{\text{fuse}}, y)
+ \sum_{i=1}^{M}
\mathcal{L}_{CE}(\mathbf{s}_{\text{enc}_i}, y)
\right] \\
& + \lambda_{hand}\mathcal{L}_{L1}(\hat{\mathbf{h}}, \mathbf{h})
\end{aligned}
\end{equation}

Here, $\mathcal{L}_{CE}$ denotes the Cross-Entropy with label smoothing ($\alpha_{\text{smooth}}=0.1$), $\mathcal{L}_{L1}$ denotes the Mean Absolute Error (L1) loss, and $\mathcal{L}_{SupCon}$ denotes the supervised contrastive loss formalized in \cite{supcon}.
Based on empirical tuning, the weighting hyperparameters were set to $\lambda_{ce}\!=\!3.0$, $\lambda_{supcon}\!=\!1.5$, $\lambda_{aux}\!=\!1.0$, and $\lambda_{hand}\!=\!2.0$.

\subsubsection{Training Hyperparameters}
EMG-CrossFormer was trained using the LAMB optimizer with default parameters ($\beta_{1} = 0.9$, $\beta_{2} = 0.999$, no weight decay) \cite{Lamb}.
LAMB was preferred over the standard ADAM optimizer \cite{adam}, as it provided greater training stability across all investigated models.
Gradient clipping with maximum norm $0.1$ was applied to further stabilize training.
A standard maximum norm value of $1.0$ was also tested but  yielded worse results.
The batch size was set to $128$. The initial learning rate was set to $5.0\cdot10^{-4}$ for the unimodal backbones and $5.0\cdot10^{-5}$ for the fusion and decoder modules.
An exponential scheduler with $\gamma = 0.99$ was used to decrease the learning rate after each epoch.
All models were trained using mixed precision to accelerate training, reduce memory usage, and better reflect potential real-world deployment on embedded devices.

The number of epochs was set to $200$ for sEMG-only training and to $100$ for multimodal approaches, as multimodal training showed faster convergence.
Early stopping was not adopted, since creating a separate validation set from the training gestures excessively reduced the number of available training samples.
The custom training loss previously described was used to provide deep supervision during training.

All other selected deep learning models (Shallow CNN, ResNet-1D, MKCNN, and NKDFF) were trained using the same set of training hyperparameters.
However, unlike EMG-CrossFormer, the learning rate was set to $5.0\cdot10^{-4}$ for the entire network, and categorical cross-entropy was used as the training loss.
For the machine learning models, hyperparameter tuning was performed using 4-fold cross-validation on the training repetitions, followed by refitting on the entire training set using the optimal hyperparameters.

Handcrafted features were standardized using a StandardScaler fitted on the training data and incorporated into the hyperparameter search pipeline.

\subsubsection{Performance Evaluation and Statistical Analysis}
\label{subsec:perf}
Model performance was evaluated using balanced accuracy, to account for class imbalance.
Additional evaluation metrics, including the F1-score and Cohen’s kappa, are reported in the Supplementary Materials.
Pairwise model comparisons were performed using the Wilcoxon signed-rank test \cite{wilcoxon} applied to subject-level performance estimates.
This non-parametric paired test matches the intra-subject design: model comparisons are computed within subjects, while the resulting subject-level differences are treated as independent observations across subjects.
$p$-values were corrected for multiple comparisons using the Benjamini-Hochberg method \cite{fdr}.
In addition to statistical significance, the mean paired improvement and its uncertainty, estimated by non-parametric bootstrap resampling, are reported to quantify the magnitude and practical relevance of performance differences.
\section{Results}
\label{sec:results}

This section summarizes the results of nearly $3,\!000$ training runs across different models, datasets, window lengths, and number of input modalities.
In particular:
\WarningsOff
\begin{enumerate}[$\bullet$]
    \item \Autoref{res:onemode} presents the comparison of models in the sEMG-only setting, showing that $\text{EMG-CF}_{\text{2D}}$ performs better than the other considered models across DB2, DB3 and DB7 for all window lengths.
    \item \Autoref{res:twomode} presents the performance gain obtained by adding accelerometer signals as a second input modality.
    \item \Autoref{res:threemode} evaluates EMG-CrossFormer performance on DB10, where gaze data are incorporated as a third input modality.
    \item \Autoref{res:compute} compares the computational cost, memory footprint, and inference latency of different EMG-CrossFormer variants.
\end{enumerate}
\WarningsOn
For each configuration (dataset, window length, and number of input modalities), the description of the results focuses on the mean paired balanced accuracy difference between the best-performing EMG-CrossFormer variant and the strongest competing model listed in \autoref{subsec:modelcomp}.
Complete model-level statistical analyses for unimodal and multimodal settings, together with additional performance metrics, are provided in the Supplementary Materials.

%------------------------------------ sEMG-only ---------------------------------------%

\subsection{Single modality: sEMG-only input}
\label{res:onemode}

\autoref{tab:unimodal} summarizes the results of the selected models in the unimodal-input setting (sEMG-only).
$\text{EMG-CF}_{\text{2D}}$ achieves the highest mean balanced accuracy for every dataset and window length, with consistent but modest improvements over the strongest competing model.
The closest competitors are RF and MKCNN, the latter providing the 2D multi-scale backbone used in $\text{EMG-CF}_{\text{2D}}$.
By contrast, $\text{EMG-CF}_{\text{1D}}$ performs below the 2D variant, suggesting that explicit spatio-temporal feature extraction is beneficial when only sEMG signals are available.

On DB2, $\text{EMG-CF}_{\text{2D}}$ surpasses RF by $0.79$ percentage points (pp) at $200$~ms ($\mathrm{CI}=[0.23, 1.31]$, $p_{\mathrm{FDR}}~=~0.003$), and MKCNN by $1.44$~pp at $150$~ms ($\mathrm{CI}=[0.97, 1.85]$, $p_{\mathrm{FDR}}<0.001$) and $1.50$~pp at $100$~ms ($\mathrm{CI}\!=\![-0.16, 2.42]$, n.s.).
On DB3, $\text{EMG-CF}_{\text{2D}}$ surpasses RF by $1.46$~pp at $200$~ms ($\mathrm{CI}=[-0.20, 3.11]$, n.s.) and $1.44$~pp at $150$~ms ($\mathrm{CI}=[-0.67, 3.35]$, n.s.), and MKCNN by $1.28$~pp at $100$~ms ($\mathrm{CI}=[-0.16, 2.42]$, n.s.).
On DB7, $\text{EMG-CF}_{\text{2D}}$ surpasses RF by $1.04$~pp at $200$~ms ($\mathrm{CI}=[0.36, 1.69]$, $p_{\mathrm{FDR}}=0.006$), and MKCNN by $1.81$~pp at $150$~ms ($\mathrm{CI}~=~[1.42, 2.20]$, $p_{\mathrm{FDR}}<0.001$) and by $2.09$~pp at $100$~ms ($\mathrm{CI}~=~[1.75, 2.45]$, $p_{\mathrm{FDR}}<0.001$).

\setlength\tabcolsep{4pt}
\begin{table*}[t]
\centering
\caption{Balanced Accuracy (\%) across models, datasets, and window lengths using only sEMG signals}
\scriptsize
\begin{tabular}{l|ccc|ccc|ccc}
\toprule
 \multirow{2}{*}{\makecell[c]{Model}}
 & \multicolumn{3}{c|}{DB2} 
 & \multicolumn{3}{c|}{DB3} 
 & \multicolumn{3}{c}{DB7} \\
 & $100$~ms  & $150$~ms  & $200$~ms  
 & $100$~ms  & $150$~ms  & $200$~ms  
 & $100$~ms  & $150$~ms  & $200$~ms \\
\midrule

SVM
  & \makecell[c]{58.29$\pm$6.87} 
  & \makecell[c]{60.34$\pm$7.11} 
  & \makecell[c]{62.05$\pm$7.37} 
  
  & \makecell[c]{39.98$\pm$7.01} 
  & \makecell[c]{42.22$\pm$7.10} 
  & \makecell[c]{43.66$\pm$7.24} 
  
  & \makecell[c]{64.91$\pm$6.38} 
  & \makecell[c]{67.46$\pm$6.51} 
  & \makecell[c]{69.49$\pm$6.65} 
  
  \\ 

 \makecell[l]{Random Forest}
  & \makecell[c]{67.26$\pm$7.06} 
  & \makecell[c]{69.86$\pm$6.86} 
  & \makecell[c]{71.55$\pm$6.74} 
  
  & \makecell[c]{46.51$\pm$8.13} 
  & \makecell[c]{49.17$\pm$7.75} 
  & \makecell[c]{51.02$\pm$7.76} 
  
  & \makecell[c]{73.70$\pm$5.88} 
  & \makecell[c]{76.48$\pm$5.64} 
  & \makecell[c]{78.11$\pm$5.51} 
  
  \\ 

 ShallowCNN
  & \makecell[c]{63.49$\pm$5.93} 
  & \makecell[c]{62.38$\pm$5.82} 
  & \makecell[c]{59.38$\pm$5.52} 
  
  & \makecell[c]{43.21$\pm$7.28} 
  & \makecell[c]{42.44$\pm$7.04} 
  & \makecell[c]{40.89$\pm$6.76} 
  
  & \makecell[c]{69.16$\pm$6.51} 
  & \makecell[c]{67.75$\pm$6.63} 
  & \makecell[c]{65.98$\pm$6.67} 
  
  \\ 

 ResNet-1D
  & \makecell[c]{62.68$\pm$6.82} 
  & \makecell[c]{63.13$\pm$6.26} 
  & \makecell[c]{63.99$\pm$6.82} 
  
  & \makecell[c]{41.63$\pm$9.72} 
  & \makecell[c]{43.31$\pm$8.18} 
  & \makecell[c]{44.03$\pm$9.89} 
  
  & \makecell[c]{70.02$\pm$7.10} 
  & \makecell[c]{70.64$\pm$6.42} 
  & \makecell[c]{71.92$\pm$6.63} 
  
  \\ 

 MKCNN
  & \makecell[c]{68.62$\pm$6.49} 
  & \makecell[c]{70.07$\pm$6.17} 
  & \makecell[c]{70.87$\pm$5.95} 
  
  & \makecell[c]{47.53$\pm$8.87} 
  & \makecell[c]{49.13$\pm$8.55} 
  & \makecell[c]{50.85$\pm$8.31} 
  
  & \makecell[c]{75.11$\pm$6.28} 
  & \makecell[c]{76.55$\pm$6.34} 
  & \makecell[c]{77.44$\pm$6.45} 
  
  \\ 

 NKDFF
  & \makecell[c]{61.86$\pm$10.58} 
  & \makecell[c]{66.15$\pm$9.61} 
  & \makecell[c]{68.39$\pm$9.77} 
  
  & \makecell[c]{43.77$\pm$8.60} 
  & \makecell[c]{44.37$\pm$14.61} 
  & \makecell[c]{47.04$\pm$12.83} 
  
  & \makecell[c]{65.64$\pm$10.16} 
  & \makecell[c]{71.08$\pm$9.69} 
  & \makecell[c]{66.93$\pm$14.94} 
  
  \\ 
  \midrule 

 $\text{EMG-CF}_{\text{1D}}$
  & \makecell[c]{66.91$\pm$7.75} 
  & \makecell[c]{68.62$\pm$7.98} 
  & \makecell[c]{70.24$\pm$7.46} 
  
  & \makecell[c]{46.64$\pm$9.74} 
  & \makecell[c]{48.14$\pm$11.48} 
  & \makecell[c]{51.32$\pm$9.73} 
  
  & \makecell[c]{74.92$\pm$5.94} 
  & \makecell[c]{76.98$\pm$6.46} 
  & \makecell[c]{77.38$\pm$6.66} 
  
  \\ 

 $\text{EMG-CF}_{\text{2D}}$
  & \makecell[c]{\textbf{70.12$\pm$6.69}} 
  & \makecell[c]{\textbf{71.51$\pm$6.28}} 
  & \makecell[c]{\textbf{72.33$\pm$6.13}} 
  
  & \makecell[c]{\textbf{48.80$\pm$9.65}} 
  & \makecell[c]{\textbf{50.61$\pm$9.80}} 
  & \makecell[c]{\textbf{52.48$\pm$8.16}} 
  
  & \makecell[c]{\textbf{77.19$\pm$6.32}} 
  & \makecell[c]{\textbf{78.36$\pm$6.33}} 
  & \makecell[c]{\textbf{79.16$\pm$6.47}} 
  
  \\ 
 
\bottomrule
\end{tabular}
\label{tab:unimodal}
\end{table*}
\setlength\tabcolsep{6pt}

%------------------------------------ sEMG + ACC ---------------------------------------%

\subsection{Two modalities: sEMG and Accelerometer inputs}
\label{res:twomode}

\setlength\tabcolsep{4pt}
\begin{table*}[t]
\centering
\caption{Balance Accuracy (\%) across models, datasets, and window lengths, using sEMG and accelerometer signals
}
\scriptsize
\begin{tabular}{l|ccc|ccc|ccc}
\toprule

 \multirow{2}{*}{Model}
 & \multicolumn{3}{c|}{DB2} 
 & \multicolumn{3}{c|}{DB3} 
 & \multicolumn{3}{c}{DB7} \\
 & $100$~ms  & $150$~ms  & $200$~ms  
 & $100$~ms  & $150$~ms  & $200$~ms  
 & $100$~ms  & $150$~ms  & $200$~ms \\
\midrule

  {SVM}
  & \makecell[c]{78.59$\pm$5.49\\(+20.31)} 
  & \makecell[c]{78.50$\pm$5.54\\(+18.16)} 
  & \makecell[c]{78.71$\pm$5.67\\(+16.66)} 
  
  & \makecell[c]{64.35$\pm$10.22\\(+24.37)} 
  & \makecell[c]{64.21$\pm$9.80\\(+21.99)} 
  & \makecell[c]{64.30$\pm$9.87\\(+20.64)} 
  
  & \makecell[c]{82.63$\pm$5.73\\(+17.73)} 
  & \makecell[c]{83.15$\pm$5.26\\(+15.68)} 
  & \makecell[c]{83.73$\pm$5.43\\(+14.24)} 
  
  \\ 

  {\makecell[l]{Random Forest}}
  & \makecell[c]{87.19$\pm$4.35\\(+19.93)} 
  & \makecell[c]{87.58$\pm$4.25\\(+17.72)} 
  & \makecell[c]{87.98$\pm$4.26\\(+16.43)} 
  
  & \makecell[c]{74.39$\pm$8.98\\(+27.88)} 
  & \makecell[c]{75.11$\pm$8.86\\(+25.94)} 
  & \makecell[c]{75.56$\pm$8.70\\(+24.55)} 
  
  & \makecell[c]{90.34$\pm$5.51\\(+16.64)} 
  & \makecell[c]{90.75$\pm$5.20\\(+14.27)} 
  & \makecell[c]{91.07$\pm$5.16\\(+12.95)} 
  
  \\ 

  {NKDFF}
  & \makecell[c]{87.78$\pm$4.03\\(+25.91)} 
  & \makecell[c]{88.24$\pm$3.71\\(+22.09)} 
  & \makecell[c]{88.61$\pm$3.79\\(+20.22)} 
  
  & \makecell[c]{73.84$\pm$8.36\\(+30.07)} 
  & \makecell[c]{76.03$\pm$8.62\\(+31.66)} 
  & \makecell[c]{76.58$\pm$8.71\\(+29.54)} 
  
  & \makecell[c]{89.65$\pm$5.18\\(+24.01)} 
  & \makecell[c]{90.35$\pm$5.71\\(+19.27)} 
  & \makecell[c]{90.61$\pm$5.36\\(+23.68)} 
  
  \\ 
 \midrule 

  {$\text{EMG-CF}_{\text{1D}}$}
  & \makecell[c]{90.16$\pm$3.55\\(+23.24)} 
  & \makecell[c]{90.43$\pm$3.25\\(+21.81)} 
  & \makecell[c]{90.34$\pm$3.52\\(+20.10)} 
  
  & \makecell[c]{79.11$\pm$7.88\\(+32.47)} 
  & \makecell[c]{78.75$\pm$7.74\\(+30.61)} 
  & \makecell[c]{78.65$\pm$9.35\\(+27.33)} 
  
  & \makecell[c]{92.39$\pm$4.10\\(+17.47)} 
  & \makecell[c]{\textbf{92.79$\pm$3.92}\\(+15.81)} 
  & \makecell[c]{\textbf{92.75$\pm$4.19}\\(+15.37)} 
  
  \\ 

  {$\text{EMG-CF}_{\text{2D}}$}
  & \makecell[c]{\textbf{90.66$\pm$3.38}\\(+20.53)} 
  & \makecell[c]{\textbf{90.61$\pm$3.05}\\(+19.10)} 
  & \makecell[c]{\textbf{90.38$\pm$3.22}\\(+18.05)} 
  
  & \makecell[c]{\textbf{80.23$\pm$6.42}\\(+31.43)} 
  & \makecell[c]{\textbf{80.40$\pm$5.95}\\(+29.78)} 
  & \makecell[c]{\textbf{79.59$\pm$5.45}\\(+27.12)} 
  
  & \makecell[c]{\textbf{92.50$\pm$4.18}\\(+15.31)} 
  & \makecell[c]{92.36$\pm$4.12\\(+14.00)} 
  & \makecell[c]{92.11$\pm$4.29\\(+12.96)} 
  
  \\

\bottomrule
\multicolumn{10}{l}{\scriptsize Parenthesized values indicate the mean balanced accuracy paired difference (pp) relative to the corresponding sEMG-only setting.}
\end{tabular}
\label{tab:twomodes}
\end{table*}
\setlength\tabcolsep{6pt}

\begin{figure*}[!t]
\centering
\includegraphics[width=\linewidth]{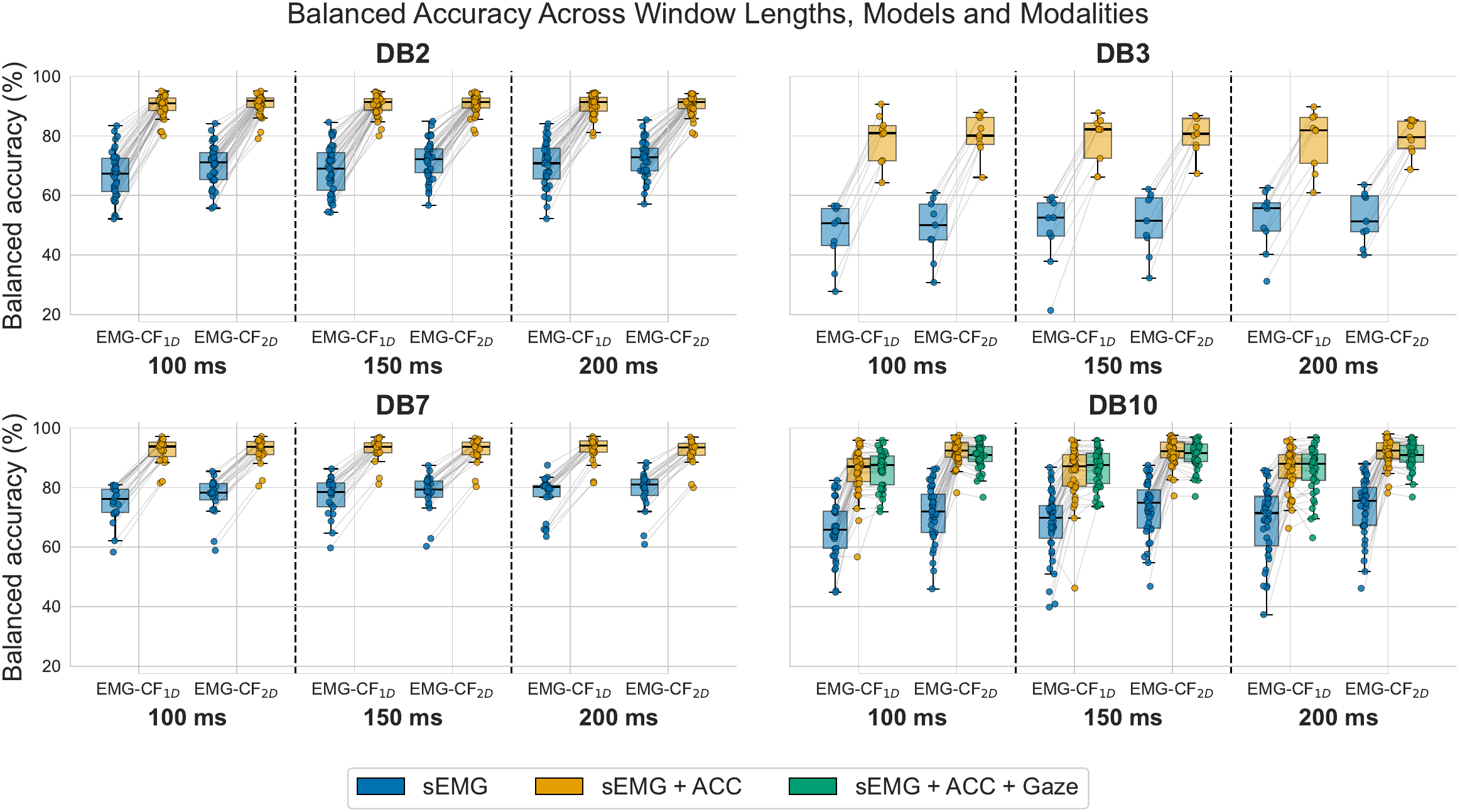}
\caption{
Hand-movement decoding performance of EMG-CrossFormer across datasets, window lengths, model variants, and input modalities. Balanced accuracy is shown for $\text{EMG-CF}_{\text{1D}}$, based on a 1D depthwise convolutional backbone, and $\text{EMG-CF}_{\text{2D}}$, based on a 2D multi-scale convolutional backbone. Results are stratified by window length ($100$~ms, $150$~ms, and $200$~ms) and input configuration: sEMG-only decoding is shown in blue, sEMG + ACC decoding is shown in orange, sEMG + ACC + Gaze in green. Each point represents one subject, and gray lines connect paired unimodal and multimodal results from the same subject. Subplots correspond to NinaPro DB2, DB3, DB7, and DB10.
}
\label{fig:results}
\end{figure*}

\autoref{tab:twomodes} summarizes the results obtained when sEMG and accelerometer signals are jointly used as input.
In this multimodal setting,  the two EMG-CrossFormer variants achieve the highest mean balanced accuracy across all datasets and window lengths.
$\text{EMG-CF}_{\text{2D}}$ ranks first on DB2 and DB3, while $\text{EMG-CF}_{\text{1D}}$ achieves the best DB7 performance at $150$~ms and $200$~ms.
In addition, EMG-CrossFormer exhibits one of the largest improvements from the unimodal to the multimodal setting, surpassed only by NKDFF; however, NKDFF always achieves a lower absolute balanced accuracy.

On DB2, $\text{EMG-CF}_{\text{2D}}$ surpasses NKDFF by $1.77$~pp at $200$~ms ($\mathrm{CI}=[1.31, 2.29]$, $p_{\mathrm{FDR}}<0.001$), $2.37$~pp at $150$~ms ($\mathrm{CI}=[1.94, 2.79]$, $p_{\mathrm{FDR}}\!<\!0.001$), and $2.88$~pp at $100$~ms ($\mathrm{CI}=[2.40, 3.36]$, $p_{\mathrm{FDR}}\!<\!0.001$).
On DB3, $\text{EMG-CF}_{\text{2D}}$ outperforms NKDFF by $3.02$~pp at $200$~ms ($\mathrm{CI}~=~[0.56, 5.64]$, n.s.), $4.37$~pp at $150$~ms ($\mathrm{CI}~=~[2.31, 6.59]$, $p_{\mathrm{FDR}}~=~0.004$), and $6.39$~pp at $100$~ms ($\mathrm{CI}~=~[4.43, 8.30]$, $p_{\mathrm{FDR}}~=~0.002$).
On DB7, $\text{EMG-CF}_{\text{1D}}$ surpasses RF by $1.05$~pp at $200$~ms ($\mathrm{CI}=[0.28, 1.88]$, n.s.) and $1.61$~pp at $150$~ms ($\mathrm{CI}~= [0.92, 2.38]$, $p_{\mathrm{FDR}}<0.001$), whereas $\text{EMG-CF}_{\text{2D}}$ surpasses RF by $2.16$~pp at $100$~ms ($\mathrm{CI}=[1.40, 2.99]$ $p_{\mathrm{FDR}}<0.001$).

%------------------------------------ sEMG + ACC + GAZE ---------------------------------------%

\subsection{Three modalities: sEMG, Accelerometers and Gaze inputs}
\label{res:threemode}

\autoref{tab:threemodes} and \autoref{fig:results} summarize the performance of EMG-CrossFormer variants on DB10 as the number of input modalities increases.
The transition from sEMG-only to multimodal decoding led to  mean paired accuracy gains of up to $+20.65$~pp, with the best accuracy obtained by $\text{EMG-CF}_{\text{2D}}$ using sEMG plus accelerometers and a $100$~ms window ($(92.06\pm3.91)\%$).
However, adding gaze does not always further improve decoding accuracy over sEMG plus accelerometers, which supports prior evidence that gaze is more informative for predicting imminent movements than for decoding executed ones.
This limited gain may also be related to the specific experimental setting adopted in the original DB10 study.

With a window length of $200$~ms, both $\text{EMG-CF}_{\text{1D}}$ and $\text{EMG-CF}_{\text{2D}}$ achieved their highest mean paired accuracy in the sEMG plus accelerometers setting, with $(85.00~\pm~7.07)\%$ ($+17.57$ pp over sEMG-only) and $(91.69\pm4.38)\%$ ($+18.20$ pp over sEMG-only), respectively.
With a window length of $150$~ms, $\text{EMG-CF}_{\text{1D}}$ achieved its highest mean paired accuracy in the three-modality setting, with $(86.40\pm6.38)\%$ ($+18.54$ pp over sEMG-only), whereas $\text{EMG-CF}_{\text{2D}}$ achieved its top performance in the sEMG plus accelerometers setting, with $(91.80\pm4.30)\%$ ($+18.85$ pp over sEMG-only).
With a window length of $100$~ms, $\text{EMG-CF}_{\text{1D}}$ achieved its highest mean paired accuracy in the three-modality setting, with $(86.19\pm5.19)\%$ ($+20.43$ pp over sEMG-only), whereas $\text{EMG-CF}_{\text{2D}}$ achieved its top performance in the sEMG plus accelerometers setting, with $(92.06\pm3.91)\%$ ($+20.65$ pp over sEMG-only).

\autoref{tab:threemodes} further stratifies EMG-CF accuracies by healthy and amputee subject groups.
Multimodal integration narrowed the performance gap between the two groups, most clearly for $\text{EMG-CF}_{\text{2D}}$: the healthy--amputee difference decreased from approximately $15$~pp with sEMG alone to $3.8$--$4.8$~pp with sEMG plus accelerometers across window lengths, although mean accuracy remained lower in amputees.

\setlength\tabcolsep{4pt}
\begin{table*}[t]
\centering
\caption{Balance Accuracy (\%) of EMG-CrossFormer variants on DB10, across window lengths and input modalities.}
\scriptsize
\begin{tabular}{l|l|ccc|ccc|ccc}
\toprule
 \multirow{3}{*}{\makecell[c]{Model}} & 
 \multirow{3}{*}{\makecell[c]{Modality$^*$}}
 & \multicolumn{9}{c}{DB10} \\
 & & \multicolumn{3}{c|}{$100$~ms} 
 & \multicolumn{3}{c|}{$150$~ms} 
 & \multicolumn{3}{c}{$200$~ms} \\
 & & Global  & Healthy  & Amputees   & Global  & Healthy  & Amputees   & Global  & Healthy  & Amputees \\
\midrule

  \multirow{3}{*}{$\text{EMG-CF}_{\text{1D}}$}
  & \scriptsize{S}
  & 65.76$\pm$9.23 & 69.16$\pm$7.60 & 57.13$\pm$7.14
  & 67.86$\pm$11.14 & 71.49$\pm$9.74 & 58.62$\pm$8.93
  & 68.41$\pm$11.94 & 72.81$\pm$9.48 & 57.22$\pm$10.12
  \\ 
  
  & \scriptsize{S + A}
  & 85.54$\pm$7.64 & 88.72$\pm$4.62 & 77.44$\pm$7.81
  & 85.11$\pm$9.09 & 88.62$\pm$5.27 & 76.18$\pm$10.55
  & \textbf{85.98$\pm$7.07} & \textbf{88.77$\pm$5.29} & \textbf{78.89$\pm$6.00}
  \\ 
  
  & \scriptsize{S + A + G}
  & \textbf{86.19$\pm$5.99} & \textbf{88.84$\pm$4.46} & \textbf{79.44$\pm$3.65}
  & \textbf{86.40$\pm$6.34} & \textbf{89.05$\pm$4.98} & \textbf{79.66$\pm$4.00}
  & 85.76$\pm$7.76 & 88.51$\pm$6.33 & 78.77$\pm$6.61
  \\ 
  
\midrule 

  \multirow{3}{*}{$\text{EMG-CF}_{\text{2D}}$}
  & \scriptsize{S}
  & 71.41$\pm$9.44 & 75.58$\pm$6.54 & 60.78$\pm$7.08
  & 72.95$\pm$9.37 & 77.08$\pm$6.49 & 62.43$\pm$7.06
  & 73.49$\pm$9.71 & 77.79$\pm$6.31 & 62.54$\pm$8.11
  \\ 
  
  & \scriptsize{S + A}
  & \textbf{92.06$\pm$3.91} & \textbf{93.12$\pm$3.70} & \textbf{89.34$\pm$3.02}
  & \textbf{91.80$\pm$4.30} & \textbf{93.08$\pm$3.89} & \textbf{88.54$\pm$3.47}
  & \textbf{91.70$\pm$4.38} & \textbf{93.05$\pm$3.88} & \textbf{88.25$\pm$3.65}
  \\ 
  
  & \scriptsize{S + A + G}
  & 90.65$\pm$4.40 & 91.92$\pm$3.91 & 87.40$\pm$3.89
  & 90.86$\pm$4.52 & 92.21$\pm$4.05 & 87.43$\pm$3.78
  & 90.86$\pm$4.43 & 92.22$\pm$3.93 & 87.38$\pm$3.64
  \\ 

\bottomrule
\multicolumn{11}{l}{$^* \text{S stands for sEMG, A stands for Accelerometer, G stands for Gaze}$}
\end{tabular}
\label{tab:threemodes}
\end{table*}
\setlength\tabcolsep{6pt}

\subsection{Computational analysis}
\label{res:compute}

\autoref{table:compute-perf} summarizes the computational cost, memory footprint, and latency of both $\text{EMG-CF}_{\text{1D}}$ and $\text{EMG-CF}_{\text{2D}}$ under unimodal and multimodal configurations.
Performance metrics were benchmarked on a single NVIDIA A30 GPU using CUDA 12.2 and PyTorch 2.12.0+cu126.
Reported values correspond to the average profiling statistics collected over $1000$ independent inference runs with a batch size of one.
Results are intended as reference benchmarks, as embedded processors used in prosthetic systems generally provide significantly lower computational capabilities than the evaluation hardware.

The results demonstrate that $\text{EMG-CF}_{\text{1D}}$ achieves the best trade-off between predictive performance and inference latency, particularly in the multimodal setting.
First, the multimodal $\text{EMG-CF}_{\text{1D}}$ model requires $5.8\times$ fewer FLOPs than its 2D counterpart.
This reduction is associated to the lightweight and efficient design of the 1D depthwise convolutional backbone, which scales more effectively with the number of input channels.
Second, the model's compiled version (through TorchInductor) yields an $8.98\times$ latency reduction for the 1D variant compared to the compiled 2D-backbone configuration under identical multimodal input.
$\text{EMG-CF}_{\text{1D}}$ is the only model achieving sub-millisecond inference latency ($0.77~\pm~0.10$~ms) on the tested hardware.

Although $\text{EMG-CF}_{\text{2D}}$ achieves higher classification accuracy across different datasets and window lengths, this performance gap becomes smaller in the multimodal setting.
This similarity in multimodal decoding accuracy, coupled with the lower computational overhead, supports the selection of the $\text{EMG-CF}_{\text{1D}}$ variant for real-time, resource-constrained edge applications such as myoelectric prosthetic control systems.

\setlength\tabcolsep{2pt}
\begin{table}[t]
\centering
\caption{Computational analysis of EMG-CrossFormer Variants}
\scriptsize
\begin{tabular}{ccccccc}
\toprule
Model & \makecell[c]{Input} & \makecell[c]{Params\\(M)} & \makecell[c]{FLOPs\\(M)} & Configuration & \makecell[c]{Latency\\(ms)} & \makecell[c]{Memory\\(Mb)} \\
\midrule
    \multirow{8}{*}{\makecell{\scriptsize{$\text{EMG-CF}_{\text{1D}}$}}}
    
    & \multirow{4}{*}{\makecell{sEMG\\$[8\!\!\times\!\!200]$\\$[4\!\!\times\!\!200]$}} & \multirow{4}{*}{1.03} & \multirow{4}{*}{54.40}
    & FP32 CPU & 5.52$\pm$0.26 & - \\
    & & & & FP32 GPU & 3.27$\pm$0.09 & 16.42  \\
    & & & & FP16 GPU & 3.62$\pm$0.44 & 12.30 \\
    & & & & \makecell[c]{Compiled} & 0.74$\pm$0.06 & 8.06 \\
    \cmidrule{2-7}
     & \multirow{4}{*}{\makecell{sEMG\\$[12\!\!\times\!\!200]$\\ACC\\$[36\!\!\times\!\!200]$}} & \multirow{4}{*}{1.07} & \multirow{4}{*}{59.04}
    & FP32 CPU & 6.15$\pm$0.38 & - \\
    & & & & FP32 GPU & 3.53$\pm$0.23 & 16.73  \\
    & & & & FP16 GPU & 3.93$\pm$0.44 & 12.46 \\
    & & & & \makecell[c]{Compiled} & 0.77$\pm$0.10 & 8.37 \\

\midrule

    \multirow{8}{*}{\makecell{\scriptsize{$\text{EMG-CF}_{\text{2D}}$}}}
    
    & \multirow{4}{*}{\makecell{sEMG\\$[8\!\!\times\!\!200]$\\$[4\!\!\times\!\!200]$}} & \multirow{4}{*}{1.20} & \multirow{4}{*}{79.18}
    & FP32 CPU & 12.26$\pm$0.22 & - \\
    & & & & FP32 GPU & 6.59$\pm$0.59 & 17.97  \\
    & & & & FP16 GPU & 6.91$\pm$0.72 & 13.11 \\
    & & & & \makecell[c]{Compiled} & 5.84$\pm$0.55 & 17.98 \\
    \cmidrule{2-7}
    & \multirow{4}{*}{\makecell{sEMG\\$[12\!\!\times\!\!200]$\\ACC\\$[36\!\!\times\!\!200]$}} & \multirow{4}{*}{1.22} & \multirow{4}{*}{342.44}
    & FP32 CPU & 15.18$\pm$1.68 & - \\
    & & & & FP32 GPU & 7.55$\pm$0.64 & 19.55  \\
    & & & & FP16 GPU & 7.89$\pm$0.44 & 13.89 \\
    & & & & \makecell[c]{Compiled} & 6.92$\pm$0.47 & 19.57 \\
    
\bottomrule
\end{tabular}
\label{table:compute-perf}
\end{table}
\setlength\tabcolsep{6pt}
\section{Discussion}
\label{sec:discussion}

% decoding challenges
Decoding complex hand movements from sEMG signals remains a challenging task.
To achieve clinical and practical viability, deep learning architectures must be designed to handle the unique characteristics of this modality.
In particular, models must extract informative representations from short temporal windows to maintain low latency.
These features must effectively capture spatio-temporal relationships that characterize distinct motor patterns, overcoming the low signal-to-noise ratio intrinsic to sEMG data.

% performance of our models are still limited 
EMG-CrossFormer was developed to address several limitations of conventional unimodal convolutional architectures and to facilitate multimodal data integration in sEMG architectures.
By leveraging a 2D multi-scale backbone capable of modeling localized spatio-temporal dynamics during feature extraction, the proposed model achieved superior performance across multiple databases and temporal window configurations within an sEMG-only baseline setup.
Although performance gains are statistically significant according to FDR-corrected signed-rank tests, the mean paired accuracy gap between EMG-CrossFormer and other competitive approaches, such as RFs or MKCNN, remains modest when considering the observed variability of results.
Furthermore, sEMG-only decoding accuracy for trans-radial amputees (DB3) degrades by $20$ pp or more compared to intact individuals (DB2 and DB7).
These findings suggest that sEMG signals alone may be insufficient to reliably discriminate among a large number of partially correlated hand movements.
Consequently, integrating additional input modalities may help distinguish between similar movements by combining complementary information related to both motor intent and the physical execution of movement.
However, the effectiveness of multimodal systems strongly depends on the design of the fusion strategy, which must enable the model to fully exploit the information contained in each modality.

% How EMG-CrossFormer differs to other multimodal strategies
Most existing multimodal deep learning approaches, such as NKDFF, combine modality-specific representations through feature concatenation before the final classification stage. 
While straightforward, this strategy limits the model's ability to learn complex interactions between modalities.
In practice, concatenation may encourage the network to rely predominantly on the most informative modality while underutilizing the complementary information provided by the others.
EMG-CrossFormer addresses this limitation through a dedicated fusion module based on cross-attention mechanisms.
By integrating multimodal interactions before the transformer decoder stage, the model learns representations in which each modality can condition its features on information extracted from the others.
This design promotes the learning of richer cross-modal relationships that cannot be captured through simple feature aggregation.
Furthermore, the supervised contrastive component of the loss function described in \autoref{sec:methods} encourages modality-specific representations to be projected into a shared embedding space while preserving class separability, facilitating more effective multimodal integration.

% performance boost in multimodal configurations
Thanks to this design, EMG-CrossFormer achieves one of the highest performance improvements when switching from unimodal to a multimodal setup, even if its unimodal performance in \autoref{res:onemode} is already superior to other models.
In particular, mean paired accuracy on amputees (DB3) improves by 31.43 pp when using the shortest window length (100 ms).
This improvements reduce the observed differences in decoding accuracy between intact subjects and trans-radial amputees.
Furthermore, the accuracy gap between EMG-CrossFormer and the closest-performing model on the same dataset and window length increases from +1.27 pp in the unimodal setting (against MKCNN) to +6.37 pp (against NKDFF).
Such an increase confirms the improved ability of the model to combine information from different signals, even for amputees, who are the targets for real-world prosthetic applications.

% modality-specific contribution and practical trade-offs
Beyond the overall benefit of multimodal fusion, the choice of the additional modality also represents an important practical factor.
Among the investigated configurations, the inclusion of eye-tracking information leads to improvements in decoding performance for $\text{EMG-CF}_{\text{1D}}$, suggesting that gaze can provide discriminative cues related to motor intention.
However, this improvement comes at the cost of increased prosthetic-system complexity, since eye tracking requires dedicated hardware, calibration procedures, and integration within a wearable or clinically usable setup.
A similar trade-off has been reported for video-based modalities, whose integration can improve decoding accuracy but generally increases computational cost due to image acquisition and processing requirements \cite{cognolato2}.
Conversely, accelerometer signals provide complementary information related to the physical execution of movement with a comparatively simpler sensing configuration. 
Therefore, the selection of auxiliary modalities should consider not only decoding accuracy, but also hardware complexity, computational cost, usability, and translational feasibility for real-world prosthetic applications.

% limitations 1: interpretability
Despite these promising results, several limitations should be acknowledged.
Although EMG-CrossFormer achieves superior performance while maintaining reduced FLOPs, memory footprint, and inference latency (see \autoref{table:compute-perf}), it remains a black box model. 
In rehabilitation and other biomedical applications, interpretability is essential for improving model reliability, understanding failure scenarios, and increasing robustness to out-of-distribution samples.
The 1D depthwise convolutional encoder adopted in $\text{EMG-CF}_{\text{1D}}$ extracts channel-specific features characterizing local portions of the input signal. 
Still, the physiological meaning of these learned representations, as well as the temporal patterns that activate them, remains unclear.
Future work should investigate architectures and analysis methods that improve the interpretability of unimodal representations.
Interpretable modality-specific features would also facilitate the analysis of multimodal interactions by enabling inspection of cross-attention matrices and quantification of attention flow through transformer layers using explainable artificial intelligence (XAI) techniques such as attention rollout \cite{attflow}.

% limitation 2: benchmarking
A second limitation concerns benchmarking and reproducibility.
Although the results presented in \autoref{sec:results} demonstrate that EMG-CrossFormer consistently ranks first among the top-performing models across all investigated datasets and configurations, comparisons between studies remain challenging.
As anticipated in \autoref{sec:introduction}, differences in preprocessing pipelines, train-test splitting strategies, hyperparameter optimization procedures, and evaluation protocols can substantially influence reported performance.
Furthermore, the lack of implementation details and open repositories identified in many studies prevents the inclusion of other models in the presented benchmarking.
As the number of deep learning studies for hand gesture decoding continues to grow, there is an increasing need for standardized benchmarking frameworks that facilitate fair comparisons and fully reproducible evaluations.
The source code made openly available in this study enables an easy integration of different models in the same experimental setting described in \autoref{sec:methods}.
The community is therefore encouraged to add more models and support the design of an open benchmarking library.

% limitation 3: datasets
Finally, the experiments were conducted primarily on the NinaPro database, which represents one of the largest and most widely adopted resource for hand movement decoding.
Nevertheless, both the number of subjects and the diversity of modalities remain limited for large multimodal deep learning applications.
The acquisition and public release of larger multimodal datasets could accelerate progress in this field by enabling more comprehensive evaluations of emerging architectures.
Such datasets may also support the development of foundation models for sEMG analysis that can be efficiently adapted to potential end-users through zero-shot, few-shot, or transfer-learning strategies.

\section{Conclusion}
\label{sec:conclusion}

This work confirms that multimodal data fusion is a powerful approach for improving unimodal sEMG-based hand gesture recognition while requiring only a limited increase in computational workload.
To this end, EMG-CrossFormer is proposed as an end-to-end hybrid convolu-tional-transformer model for the seamless integration of multiple signals.
EMG-CrossFormer combines representations from an arbitrary number of unimodal encoders through cascaded cross-attention layers, and decodes the fused representations using learnable gesture queries.
The architecture was designed to allow researchers to easily customize its modules while maintaining a computational footprint suitable for embedded deployment.
Unimodal backbones may differ across modalities, and the number of modalities can vary depending on the target application.
EMG-CrossFormer was evaluated on four NinaPro databases (DB2, DB3, DB7, and DB10) and benchmarked against six state-of-the-art models.
When trained using only sEMG signals, EMG-CrossFormer with a 2D multi-scale backbone surpassed other competing models in decoding accuracy, although the improvements were modest.
In the multimodal setting, EMG-CrossFormer's decoding accuracy increased significantly for both intact and amputee subjects, maintaining the best overall performance and further widening the accuracy gap relative to the other models.
These findings demonstrate that modern deep learning architectures designed for multimodal data integration can effectively leverage complementary physiological information to improve hand gesture decoding performance.
However, the development of solutions that can be safely and reliably deployed in real-world assistive devices remains an open challenge for the scientific community.

\printcredits

\section*{Declaration of competing interest}
The authors declare that they have no known competing financial interests or personal relationships that could have appeared to influence the work reported in this paper.

\section*{Code and data availability}
The code used to produce both results and figures is openly available at \href{https://github.com/DeepPNCLab/emg-crossformer}{https://github.com/DeepPNCLab/emg-crossformer}.
All data that support the findings of this study are openly available at \href{https://ninapro.hevs.ch}{https://ninapro.hevs.ch}.

\section*{Acknowledgment}
This document is the result of the research project funded by the European Unions Horizon Europe research and innovation programme under Grant agreement no 101137074 - HEREDITARY.

\bibliographystyle{elsarticle-num}
\bibliography{bibliography_abbreviated.bib}

\setcounter{table}{0}
\setcounter{figure}{0}
\appendix
\onecolumn

\section{Supplementary Materials}

This document provides supplementary materials for the research work Multimodal Surface EMG Hand Gesture Recognition Using Query-Based Transformers for Prosthetic Control.
Specifically, it complements the methodology described in Section II of the main text and further supports the results presented in Section III.

\subsection{Further details on model architecture}

\begin{figure}[!ht]
\centering
\includegraphics[width=\linewidth]{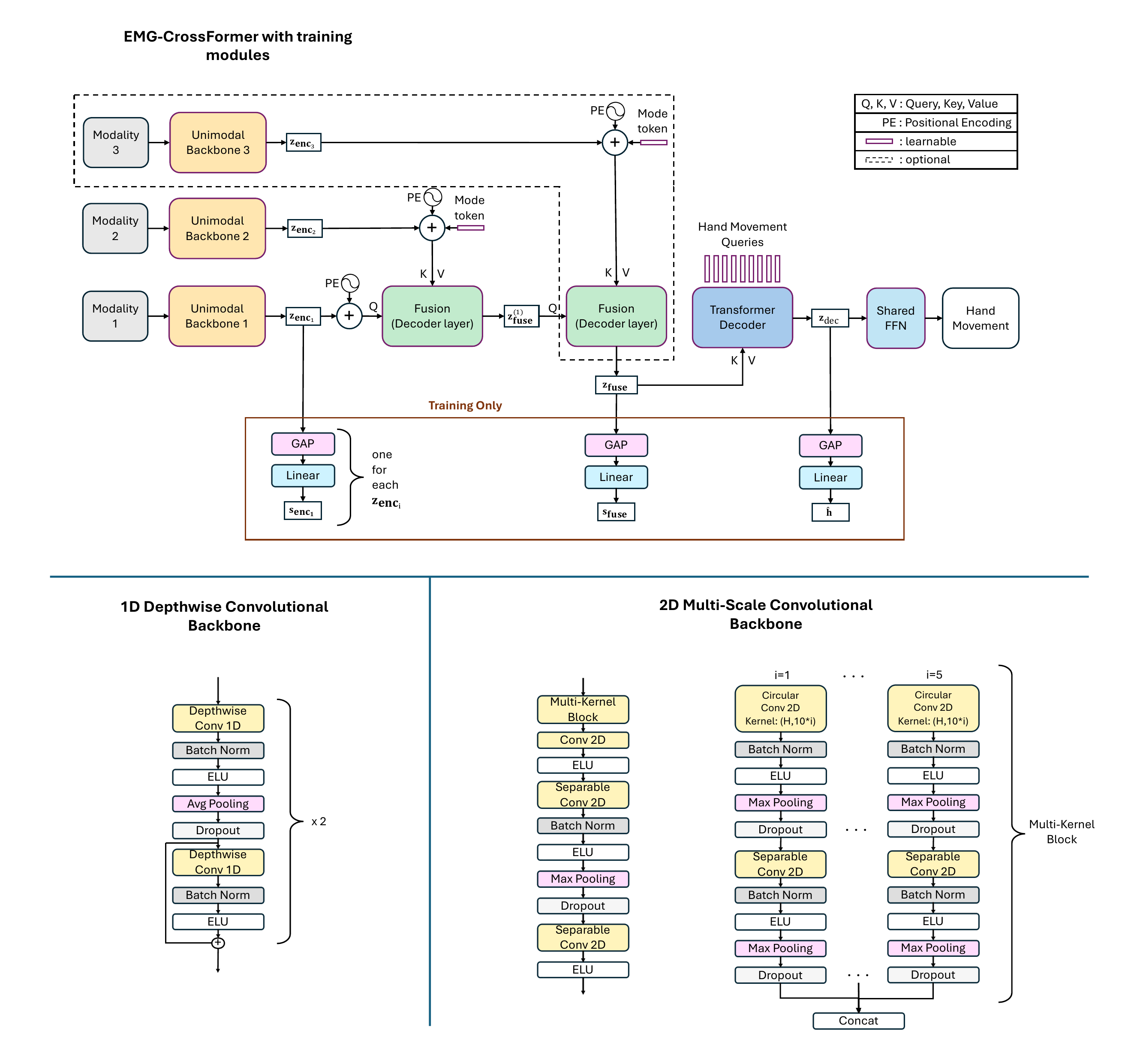}
\caption{
Detailed representation of the EMG-CrossFormer architecture.
The model combines representations ($\mathbf{z}_{\text{enc}_i}$) from an arbitrary number of unimodal backbones through cascaded fusion layers (transformer decoder layers).
The transformer decoder block decodes the fused representations ($\mathbf{z}_{\text{fuse}}$) using learnable hand-movement queries.
A final shared FFN outputs hand-movement predictions using the decoded representations ($\mathbf{z}_{\text{dec}}$).
During training, auxiliary predictions were generated from intermediate representations by applying Global Average Pooling (GAP), followed by a linear projection layer.
}
\label{fig:emgcrossformer_supplementary}
\end{figure}

\autoref{fig:emgcrossformer_supplementary} expands on the EMG-CrossFormer architecture, reporting the structure of the network modules used to train the model, the unimodal backbones, and the initialization hyperparameters. 

\clearpage
\noindent \textbf{1D Depthwise Convolutional Backbone}.
This backbone consists of two stacked depthwise convolutional blocks.
Each block applies a depthwise 1D convolution, batch normalization, ELU activation, average pooling, and dropout, followed by a second depthwise convolution, batch normalization, and ELU activation.
A residual connection adds the feature representation immediately after dropout to the output of the second convolutional stage.
The output of the second block is projected to the EMG-CrossFormer embedding dimension through a two-layer FFN with hidden size 128 and ReLU activation.

\captionsetup{hypcap=false}
\begin{center}
\captionof{table}{Hyperparameters of the\\1D depthwise backbone.}
\label{tab:supp1}
\begin{tabular}{lc}
\toprule
Parameter & Value \\
\midrule
First depthwise multiplier & 2 \\
Second depthwise multiplier & 1 \\
Kernel length & 41 \\
Average pooling kernel size & 2 \\
Average pooling stride & 2 \\
Dropout rate & 0.025 \\
ELU $\alpha$ & 0.1 \\
BatchNorm momentum & 0.25 \\
\bottomrule
\end{tabular}
\end{center}
\captionsetup{hypcap=true}

\vspace{1em}
\noindent \textbf{2D Multi-Scale Convolutional Backbone}.

\noindent This backbone combines parallel 2D convolutional branches that extract temporal features at different scales (referred to as the \emph{multi-kernel block}), followed by a shared feature-mixing branch composed of separable 2D convolutions.

\noindent As schematized in \autoref{fig:emgcrossformer_supplementary}, the input signal is first reshaped into a single-channel pseudo-image by introducing a channel dimension.
The resulting representation is processed by a multi-kernel block consisting of five parallel branches.
Each branch applies a 2D convolution with circular padding to account for the spatial arrangement of forearm electrodes, followed by batch normalization, ELU activation, max pooling, dropout, a separable 2D convolution, and a second sequence of batch normalization, ELU activation, max pooling, and dropout.

The outputs of the five branches are concatenated and passed to a shared feature-mixing branch composed of a $1\times1$ convolution, ELU activation, a separable 2D convolution followed by batch normalization and ELU activation, adaptive max pooling, dropout, a final separable 2D convolution, and ELU activation.
The final representation is obtained by flattening the last two dimensions and projecting it to the EMG-CrossFormer embedding dimension through a linear layer.

Initialization hyperparameters are provided in the openly available source code.
The parameters of the multi-kernel block and shared feature-mixing branch are reported in Tables~\ref{tab:multikernel_params} and \ref{tab:mixing_params}, respectively.

\captionsetup{hypcap=false}
\begin{center}
\captionof{table}{Hyperparameters of the multi-kernel block.}
\label{tab:multikernel_params}
\begin{tabular}{ll}
\toprule
Parameter & Value \\
\midrule
Number of parallel branches & 5 \\
Output channels (first convolution) & 32 \\
Kernel size & $C \times (10i)$, $i \in \{1,2,3,4,5\}$ \\
Channel dimension $C$ & 3 (sEMG), 9 (accelerometers), 1 (gaze) \\
Circular padding & 1 (sEMG), 3 (accelerometers), 0 (gaze) \\
Circular padding application & Forearm-channel dimension only \\
First max pooling kernel size & $[1,20]$ \\
First max pooling stride & 1 \\
Separable convolution output channels & 64 \\
Separable convolution depthwise multiplier & 1 \\
Separable convolution kernel size & $3\times3$ \\
Dropout rate & 0.2 \\
Second max pooling kernel size & $[2,2]$ \\
Second max pooling stride & 1 \\
\bottomrule
\end{tabular}
\end{center}
\captionsetup{hypcap=true}

\clearpage
\captionsetup{hypcap=false}
\begin{center}
\captionof{table}{Hyperparameters of the shared feature-mixing branch.}
\label{tab:mixing_params}
\begin{tabular}{ll}
\toprule
Parameter & Value \\
\midrule
$1\times1$ convolution output channels & 128 \\
Separable convolution output channels & 128 \\
Separable convolution depthwise multiplier & 1 \\
Separable convolution kernel size & $3\times3$ \\
Adaptive max pooling output size & $(5,2)$ \\
Dropout rate & 0.2 \\
\bottomrule
\end{tabular}
\end{center}
\captionsetup{hypcap=true}

\noindent \textbf{Transformer Decoder}.

\noindent The decoder follows the default PyTorch implementation and consists of four decoder layers.
Each layer uses an embedding dimension of 128, eight attention heads, a feedforward hidden dimension of 128, no dropout, and a SwiGLU activation function.

\captionsetup{hypcap=false}
\begin{center}
\captionof{table}{Transformer decoder hyperparameters.}
\label{tab:decoder_params}
\begin{tabular}{ll}
\toprule
Parameter & Value \\
\midrule
Number of decoder layers & 4 \\
Embedding dimension & 128 \\
Number of attention heads & 8 \\
Feedforward hidden dimension & 128 \\
Dropout rate & 0.0 \\
Activation function & SwiGLU \\
\bottomrule
\end{tabular}
\end{center}
\captionsetup{hypcap=true}

\vspace{1em}
\noindent \textbf{Train-only Layers}.

\noindent During training, auxiliary predictions were generated from intermediate representations by applying Global Average Pooling (GAP), followed by a linear projection layer.

\subsection{Further details on the Machine Learning pipeline}

\noindent A set of handcrafted features commonly used in prior work was extracted and used to train the machine learning models. Specifically, the following 29 features were extracted from each channel:

\begin{itemize}
    \item Root Mean Square.
    \item Mean Absolute Value.
    \item Waveform Length: cumulative length of the waveform, computed as the sum of the absolute differences between consecutive samples.
    \item Zero Crossings: number of sign changes in the signal.
    \item Slope Sign Changes: number of sign changes in the first derivative of the signal.
    \item Histogram Features: counts of samples that fall into each bin. The signal is quantized into 20 bins defined over the range $[-3\sigma, 3\sigma]$, producing 20 features per channel.
    \item Marginal Discrete Wavelet Transform: computed using a Daubechies-7 (\texttt{db7}) wavelet decomposition with 3 levels, producing 4 features per channel.
\end{itemize}

\noindent The implementation of the feature extraction procedure is available in the open-source code repository.

\noindent Machine learning hyperparameter tuning was performed using 4-fold cross-validation on the training repetitions, followed by refitting on the entire training set using the identified optimal set of hyperparameters. 
\autoref{tab:mlhyper} lists all optimized hyperparameters and the value grid.
Due to the long training time required to test all hyperparameter combinations, a preliminary screening was performed to identify the most suitable sub-grid and reduce the overall training time.

\clearpage
\captionsetup{hypcap=false}
\begin{center}
\captionof{table}{Machine Learning model hyperparameters.}
\label{tab:mlhyper}
\begin{tabular}{lll}
\toprule
Model & Parameter & Values \\
\midrule
\multirow{3}{*}{Random Forest}
& Number of estimators & $100, 200, 500$ \\
& maximum depth & $\text{None}, 10, 20, 30$ \\
& minimum samples to split a node & $2, 5, 10$ \\
\midrule
\multirow{3}{*}{SVM}
& kernel & poly, rbf \\
& C & $0.0001, 0.001, 0.01, 0.1, 1.0$ \\
& $\gamma$ & $0.0001, 0.001, 0.01, 0.1, 1.0$ \\
\bottomrule
\end{tabular}
\end{center}
\captionsetup{hypcap=true}

\vspace{1cm}
\subsection{Further result visualization and statistical analysis}
\subsubsection{sEMG-only setting}
\begin{figure*}[!ht]
\centering
\includegraphics[width=\linewidth]{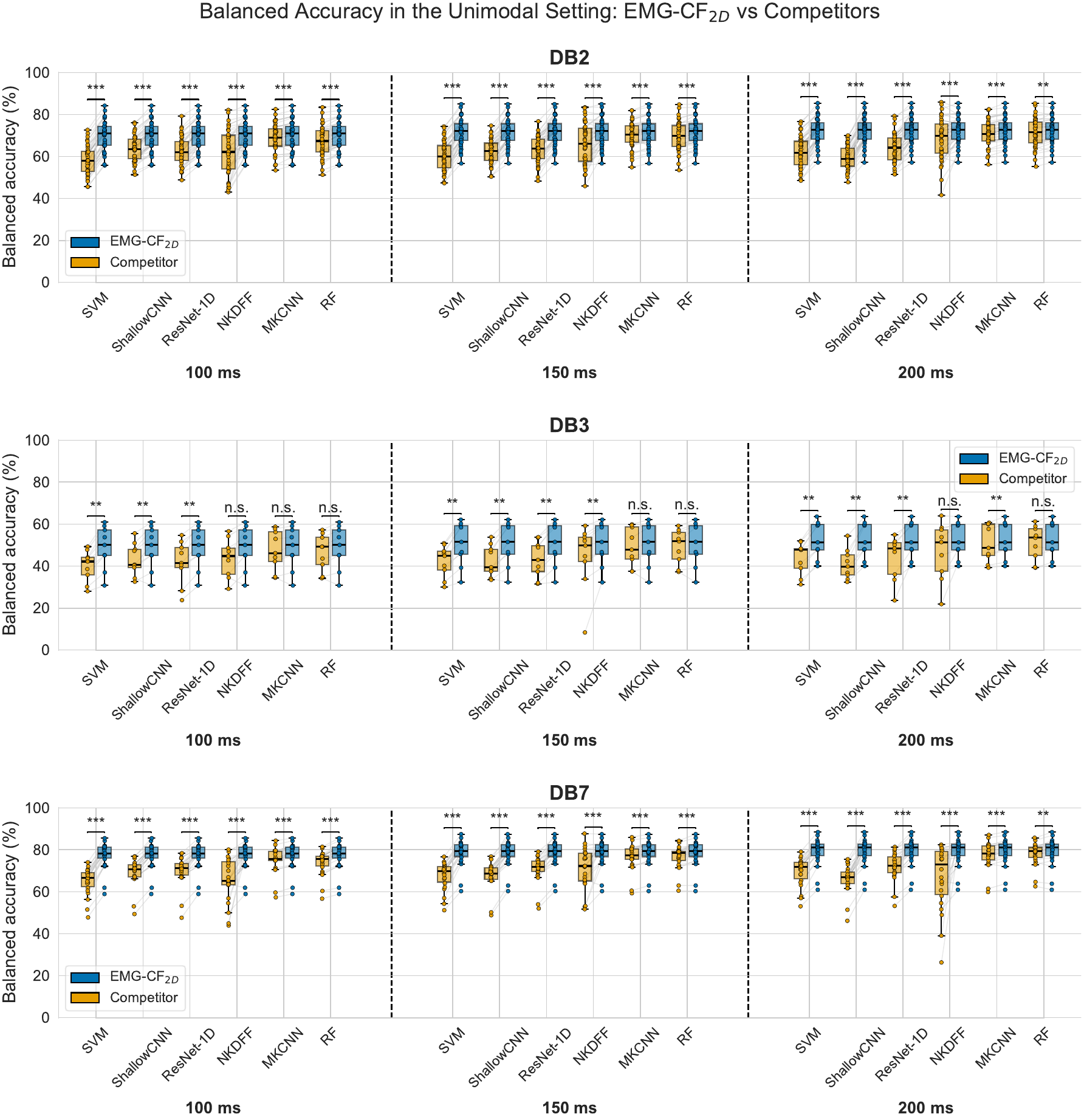}
\caption{
sEMG-only paired comparisons between EMG-CF$_{\text{2D}}$ and the competing models for NinaPro DB2, DB3, and DB7 across 100, 150, and 200~ms windows. Boxplots summarize the subject-level balanced accuracies; points represent individual subjects, and gray lines connect paired observations. Asterisks indicate one-sided paired Wilcoxon signed-rank tests adjusted across the 108 unimodal comparisons using the Benjamini-Hochberg FDR procedure: $^{***}p_{\mathrm{FDR}}<0.001$ and $^{**}p_{\mathrm{FDR}}<0.01$. Comparisons with $p_{\mathrm{FDR}}\geq0.01$ are denoted n.s.
}
\label{fig:SI_unimodal_2D}
\end{figure*}

\begin{figure*}[!ht]
\centering
\includegraphics[width=\linewidth]{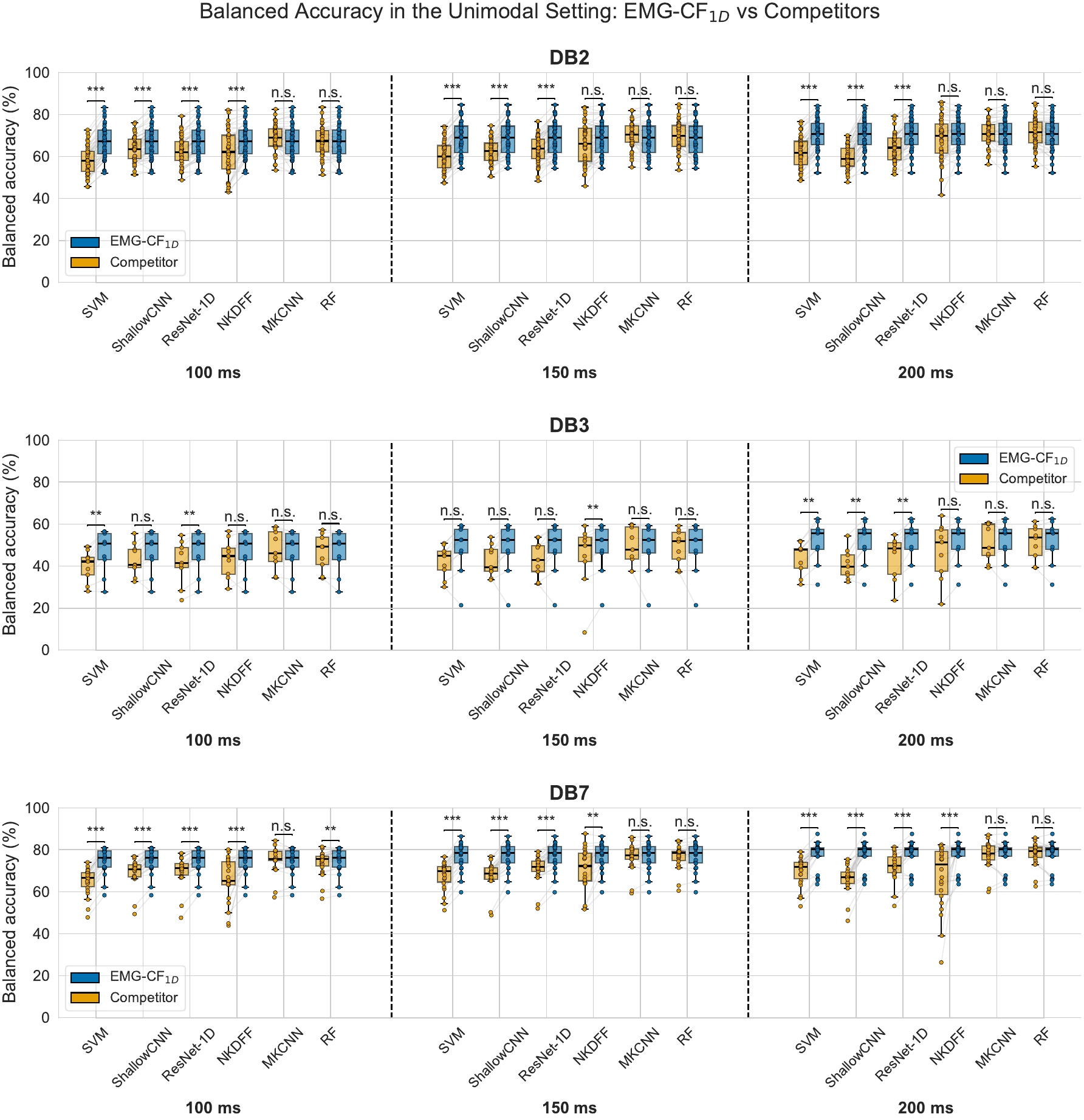}
\caption{sEMG-only paired comparisons between EMG-CF$_{\text{1D}}$ and the competing models for NinaPro DB2, DB3, and DB7 across 100, 150, and 200~ms windows. Boxplots summarize the subject-level balanced accuracies; points represent individual subjects, and gray lines connect paired observations. Asterisks indicate one-sided paired Wilcoxon signed-rank tests adjusted across the 108 unimodal comparisons using the Benjamini--Hochberg FDR procedure: $^{***}p_{\mathrm{FDR}}<0.001$ and $^{**}p_{\mathrm{FDR}}<0.01$. Comparisons with $p_{\mathrm{FDR}}\geq0.01$ are denoted n.s.
}
\label{fig:SI_unimodal_1D}
\end{figure*}

Each EMG-CrossFormer variant was compared with each competing model using subject-level balanced accuracies paired by subject
(\autoref{fig:SI_unimodal_2D} and \autoref{fig:SI_unimodal_1D}).
Comparisons were performed separately for each database and window length, yielding 108 tests:
3 databases $\times$ 3 window lengths $\times$ 2 EMG-CrossFormer variants $\times$ 6 competing models.
For each comparison, we computed the mean paired difference in balanced accuracy (EMG-CF minus competitor, in percentage points) and its 95\% bootstrap confidence interval by resampling the subject-level differences 10,000 times.
Statistical significance was assessed using a one-sided paired Wilcoxon signed-rank test, with the alternative hypothesis that EMG-CrossFormer achieved higher balanced accuracy.
The resulting 108 p-values were jointly adjusted using the Benjamini--Hochberg FDR procedure, with adjusted $p<0.01$ considered significant.

EMG-CF$_{\text{2D}}$ showed the most consistent advantage (\autoref{fig:SI_unimodal_2D}). In DB2 and DB7, it significantly outperformed all competing models at all window lengths.
In DB3, it significantly outperformed SVM, ShallowCNN, and ResNet-1D at all window lengths. Comparisons with NKDFF and MKCNN reached significance at only one window length each, while no comparison with Random Forest was significant.

EMG-CF$_{\text{1D}}$ showed a weaker and less consistent pattern
(\autoref{fig:SI_unimodal_1D}).
Significant improvements over SVM, ShallowCNN, and ResNet-1D were frequent in DB2 and DB7, but less consistent in DB3 and against NKDFF.
EMG-CF$_{\text{1D}}$ did not significantly outperform MKCNN in any configuration and significantly outperformed Random Forest only in DB7 at 100~ms.

\vspace{1em}
\subsubsection{sEMG + ACC setting}

We next tested whether EMG-CrossFormer retained its advantage when inertial information was added to the sEMG input. In this multimodal setting, each EMG-CrossFormer variant was compared with the competing models available for sEMG + ACC decoding: SVM, NKDFF, and Random Forest. Comparisons were performed separately for each database and window length, yielding 54 tests: 3 databases $\times$ 3 window lengths $\times$ 2 EMG-CrossFormer variants $\times$ 3 competing models. For each comparison, we computed the mean paired difference in balanced accuracy (EMG-CF minus competitor, in percentage points) and its 95\% bootstrap confidence interval by resampling the subject-level differences 10,000 times. Statistical significance was assessed using a one-sided paired Wilcoxon signed-rank test. The resulting 54 p-values were jointly adjusted using the Benjamini--Hochberg FDR procedure, with adjusted $p<0.01$ considered significant.

\begin{figure*}[!ht]
\centering
\includegraphics[width=\linewidth]{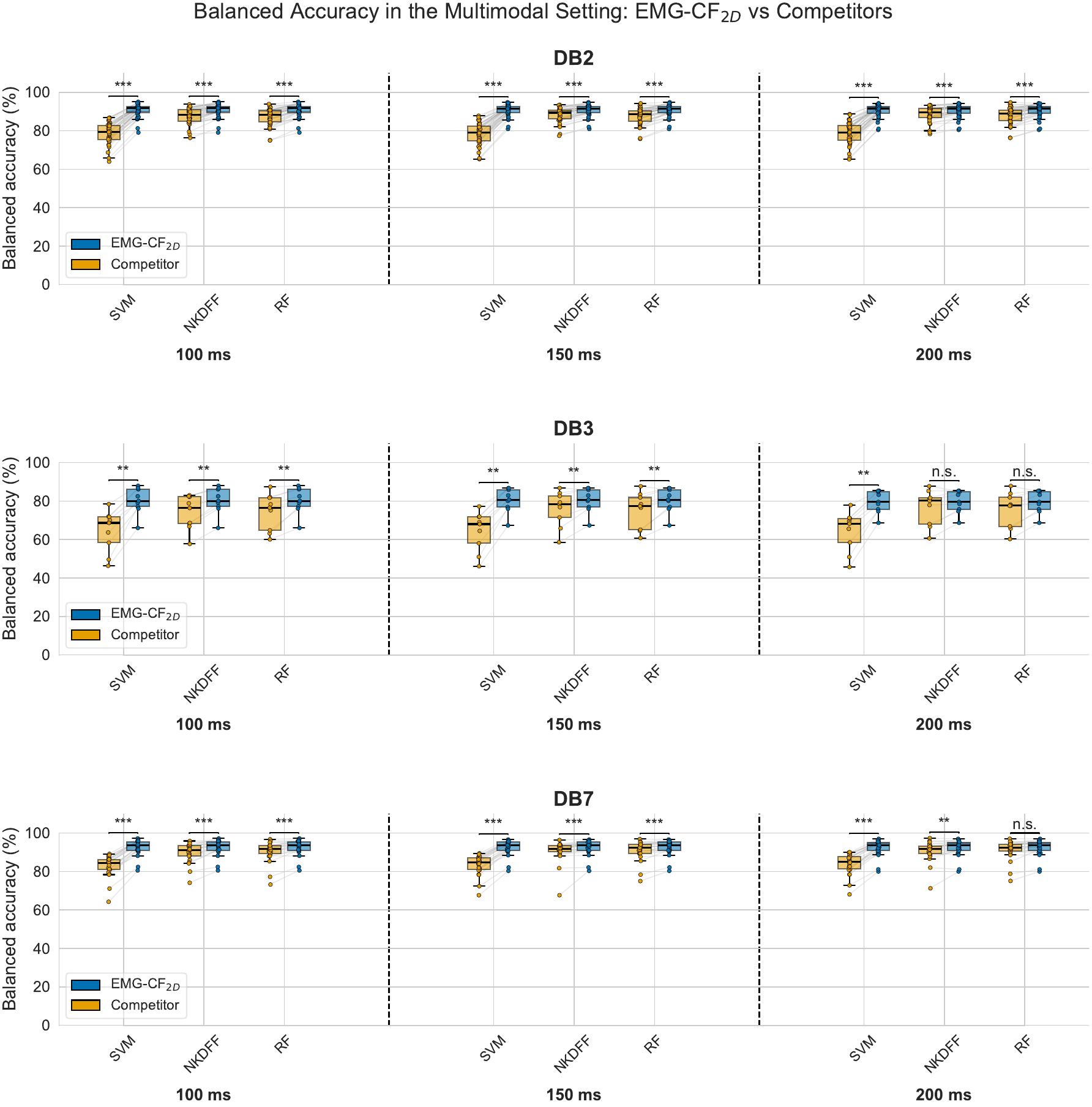}
\caption{sEMG + ACC paired comparisons between EMG-CF$_{\text{2D}}$ and the competing multimodal models for NinaPro DB2, DB3, and DB7 across 100, 150, and 200~ms windows. Boxplots summarize the subject-level balanced accuracies; points represent individual subjects, and gray lines connect paired observations. Asterisks indicate one-sided paired Wilcoxon signed-rank tests adjusted across the 54 multimodal comparisons using the Benjamini--Hochberg FDR procedure: $^{***}p_{\mathrm{FDR}}<0.001$ and $^{**}p_{\mathrm{FDR}}<0.01$. Comparisons with $p_{\mathrm{FDR}}\geq0.01$ are denoted n.s.
}
\label{fig:SI_multimodal_2D}
\end{figure*}

\begin{figure*}[!ht]
\centering
\includegraphics[width=\linewidth]{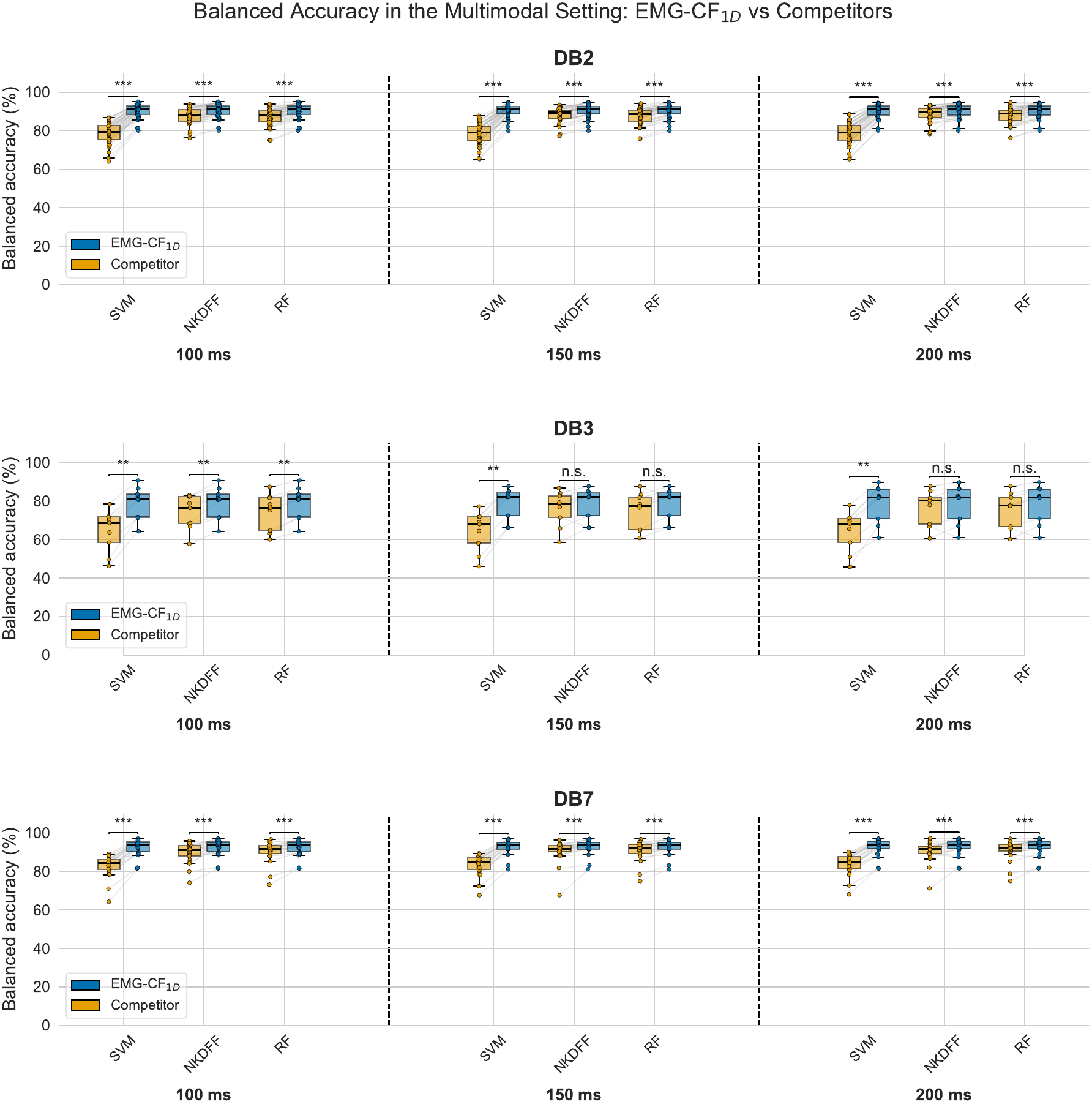}
\caption{sEMG + ACC paired comparisons between EMG-CF$_{\text{1D}}$ and the competing multimodal models for NinaPro DB2, DB3, and DB7 across 100, 150, and 200~ms windows. Boxplots summarize the subject-level balanced accuracies; points represent individual subjects, and gray lines connect paired observations. Asterisks indicate one-sided paired Wilcoxon signed-rank tests adjusted across the 54 multimodal comparisons using the Benjamini--Hochberg FDR procedure: $^{***}p_{\mathrm{FDR}}<0.001$ and $^{**}p_{\mathrm{FDR}}<0.01$. Comparisons with $p_{\mathrm{FDR}}\geq0.01$ are denoted n.s.
}
\label{fig:SI_multimodal_1D}
\end{figure*}

Both EMG-CrossFormer variants showed strong multimodal performance. EMG-CF$_{\text{1D}}$ significantly outperformed all three competing models at all window lengths in DB2 and DB7 (\autoref{fig:SI_multimodal_1D}). In DB3, it significantly outperformed all competitors at 100~ms, whereas only the comparisons with SVM remained significant at 150 and 200~ms. 
EMG-CF$_{\text{2D}}$ significantly outperformed all competing models at all window lengths in DB2 (\autoref{fig:SI_multimodal_2D}). In DB7, all comparisons were significant except that with Random Forest at 200~ms. In DB3, all comparisons were significant at 100 and 150~ms, whereas only the comparison with SVM remained significant at 200~ms.

Overall, the advantage of EMG-CrossFormer in the multimodal sEMG + ACC setting was highly consistent in DB2 and DB7 and less consistent in DB3, particularly at longer window lengths against NKDFF and Random Forest.

\clearpage
\subsection{Summary results with other metrics}
\autoref{tab:cohen} and \autoref{tab:f1} compare EMG-CrossFormer with the other models using additional metrics, namely Cohen's kappa and the F1-score. Regardless of the metric used, EMG-CrossFormer maintains superior performance, demonstrating stronger decoding capabilities than competing models.

\captionsetup{hypcap=false}
\begin{center}
\scriptsize
\captionof{table}{F1-score across models, datasets, window lengths, and number of modalities}
\begin{tabular}{l|l|ccc|ccc|ccc}
\toprule

 \multirow{2}{*}{Model} & \multirow{2}{*}{Modality}
 & \multicolumn{3}{c|}{DB2} 
 & \multicolumn{3}{c|}{DB3} 
 & \multicolumn{3}{c}{DB7} \\
 &
 & 100ms  & 150ms  & 200ms  
 & 100ms  & 150ms  & 200ms  
 & 100ms  & 150ms  & 200ms \\
\midrule

\multirow{2}{*}{SVM} 
 & \scriptsize{sEMG} 
 & $0.60\!\pm\!0.07$ & $0.62\!\pm\!0.07$ & $0.64\!\pm\!0.07$ 
  & $0.41\!\pm\!0.07$ & $0.43\!\pm\!0.07$ & $0.44\!\pm\!0.07$ 
  & $0.65\!\pm\!0.06$ & $0.68\!\pm\!0.06$ & $0.70\!\pm\!0.07$ 
  
 \\ 
 & \scriptsize{sEMG + ACC} 
 & $0.80\!\pm\!0.05$ & $0.80\!\pm\!0.05$ & $0.81\!\pm\!0.05$ 
  & $0.66\!\pm\!0.10$ & $0.66\!\pm\!0.09$ & $0.66\!\pm\!0.09$ 
  & $0.83\!\pm\!0.06$ & $0.84\!\pm\!0.05$ & $0.84\!\pm\!0.05$ 
  
 \\ 
\midrule 

\multirow{2}{*}{\makecell[l]{Random\\Forest}} 
 & \scriptsize{sEMG} 
 & $0.68\!\pm\!0.07$ & $0.70\!\pm\!0.07$ & $0.72\!\pm\!0.07$ 
  & $0.46\!\pm\!0.08$ & $0.49\!\pm\!0.08$ & $0.50\!\pm\!0.08$ 
  & $0.74\!\pm\!0.06$ & $0.76\!\pm\!0.06$ & $0.78\!\pm\!0.06$ 
  
 \\ 
 & \scriptsize{sEMG + ACC} 
 & $0.88\!\pm\!0.04$ & $0.88\!\pm\!0.04$ & $0.88\!\pm\!0.04$ 
  & $0.75\!\pm\!0.09$ & $0.76\!\pm\!0.09$ & $0.76\!\pm\!0.09$ 
  & $0.90\!\pm\!0.06$ & $0.91\!\pm\!0.05$ & $0.91\!\pm\!0.05$ 
  
 \\ 
\midrule 

\multirow{2}{*}{ShallowCNN} 
 & \scriptsize{sEMG} 
 & $0.63\!\pm\!0.06$ & $0.62\!\pm\!0.06$ & $0.59\!\pm\!0.06$ 
  & $0.42\!\pm\!0.07$ & $0.42\!\pm\!0.07$ & $0.40\!\pm\!0.06$ 
  & $0.69\!\pm\!0.07$ & $0.67\!\pm\!0.07$ & $0.66\!\pm\!0.07$ 
  
 \\ 
 & \scriptsize{sEMG + ACC} 
 & ---  & ---  & ---  
  & ---  & ---  & ---  
  & ---  & ---  & ---  
  
 \\ 
\midrule 

\multirow{2}{*}{ResNet-1D} 
 & \scriptsize{sEMG} 
 & $0.62\!\pm\!0.07$ & $0.63\!\pm\!0.06$ & $0.64\!\pm\!0.07$ 
  & $0.41\!\pm\!0.10$ & $0.42\!\pm\!0.08$ & $0.43\!\pm\!0.10$ 
  & $0.70\!\pm\!0.07$ & $0.70\!\pm\!0.07$ & $0.72\!\pm\!0.07$ 
  
 \\ 
 & \scriptsize{sEMG + ACC} 
 & ---  & ---  & ---  
  & ---  & ---  & ---  
  & ---  & ---  & ---  
  
 \\ 
\midrule 

\multirow{2}{*}{MKCNN} 
 & \scriptsize{sEMG} 
 & $0.68\!\pm\!0.07$ & $0.70\!\pm\!0.06$ & $0.70\!\pm\!0.06$ 
  & $0.47\!\pm\!0.09$ & $0.48\!\pm\!0.08$ & $0.50\!\pm\!0.08$ 
  & $0.75\!\pm\!0.06$ & $0.76\!\pm\!0.06$ & $0.77\!\pm\!0.07$ 
  
 \\ 
 & \scriptsize{sEMG + ACC} 
 & ---  & ---  & ---  
  & ---  & ---  & ---  
  & ---  & ---  & ---  
  
 \\ 
\midrule 

\multirow{2}{*}{NKDFF} 
 & \scriptsize{sEMG} 
 & $0.62\!\pm\!0.11$ & $0.66\!\pm\!0.10$ & $0.69\!\pm\!0.10$ 
  & $0.43\!\pm\!0.08$ & $0.44\!\pm\!0.15$ & $0.46\!\pm\!0.13$ 
  & $0.65\!\pm\!0.11$ & $0.71\!\pm\!0.10$ & $0.66\!\pm\!0.16$ 
  
 \\ 
 & \scriptsize{sEMG + ACC} 
 & $0.89\!\pm\!0.04$ & $0.89\!\pm\!0.04$ & $0.89\!\pm\!0.04$ 
  & $0.75\!\pm\!0.08$ & $0.77\!\pm\!0.08$ & $0.77\!\pm\!0.08$ 
  & $0.90\!\pm\!0.05$ & $0.90\!\pm\!0.06$ & $0.91\!\pm\!0.06$ 
  
 \\ 
\midrule 

\multirow{2}{*}{$\text{EMG-CF}_{\text{1D}}$} 
 & \scriptsize{sEMG} 
 & $0.67\!\pm\!0.08$ & $0.68\!\pm\!0.08$ & $0.70\!\pm\!0.08$ 
  & $0.46\!\pm\!0.10$ & $0.47\!\pm\!0.11$ & $0.50\!\pm\!0.10$ 
  & $0.75\!\pm\!0.06$ & $0.77\!\pm\!0.07$ & $0.77\!\pm\!0.07$ 
  
 \\ 
 & \scriptsize{sEMG + ACC} 
 & $0.90\!\pm\!0.04$ & $0.90\!\pm\!0.03$ & $0.90\!\pm\!0.04$ 
  & $0.80\!\pm\!0.07$ & $0.79\!\pm\!0.07$ & $0.79\!\pm\!0.09$ 
  & $\mathbf{0.92\!\pm\!0.04}$ & $\mathbf{0.93\!\pm\!0.04}$ & $\mathbf{0.93\!\pm\!0.04}$
  
 \\ 
\midrule 

\multirow{2}{*}{$\text{EMG-CF}_{\text{2D}}$} 
 & \scriptsize{sEMG} 
 &  $\mathbf{0.70\!\pm\!0.07}$ & $\mathbf{0.71\!\pm\!0.06}$ & $\mathbf{0.72\!\pm\!0.06}$ 
  & $\mathbf{0.48\!\pm\!0.10}$ & $\mathbf{0.50\!\pm\!0.10}$ & $\mathbf{0.52\!\pm\!0.08}$ 
  & $\mathbf{0.77\!\pm\!0.06}$ & $\mathbf{0.78\!\pm\!0.06}$ & $\mathbf{0.79\!\pm\!0.07}$ 
  
 \\ 
 & \scriptsize{sEMG + ACC} 
  & $\mathbf{0.91\!\pm\!0.03}$ & $\mathbf{0.91\!\pm\!0.03}$ & $\mathbf{0.90\!\pm\!0.03}$
  & $\mathbf{0.81\!\pm\!0.06}$ & $\mathbf{0.81\!\pm\!0.05}$ & $\mathbf{0.80\!\pm\!0.05}$ 
  & $\mathbf{0.92\!\pm\!0.04}$ & $0.92\!\pm\!0.04$ & $0.92\!\pm\!0.04$ 
  
 \\

\bottomrule
\end{tabular}
\label{tab:f1}
\end{center}
\captionsetup{hypcap=true}

\captionsetup{hypcap=false}
\begin{center}
\scriptsize
\captionof{table}{Cohen's Kappa across models, datasets, window lengths, and number of modalities}
\begin{tabular}{l|l|ccc|ccc|ccc}
\toprule

 \multirow{2}{*}{Model} & \multirow{2}{*}{Modality}
 & \multicolumn{3}{c|}{DB2} 
 & \multicolumn{3}{c|}{DB3} 
 & \multicolumn{3}{c}{DB7} \\
 &
 & 100ms  & 150ms  & 200ms  
 & 100ms  & 150ms  & 200ms  
 & 100ms  & 150ms  & 200ms \\
\midrule

\multirow{2}{*}{SVM} 
 & \scriptsize{sEMG} 
 & $0.55\!\pm\!0.07$ & $0.57\!\pm\!0.08$ & $0.59\!\pm\!0.08$ 
  & $0.37\!\pm\!0.07$ & $0.39\!\pm\!0.07$ & $0.40\!\pm\!0.07$ 
  & $0.63\!\pm\!0.07$ & $0.66\!\pm\!0.07$ & $0.68\!\pm\!0.08$ 
  
 \\ 
 & \scriptsize{sEMG + ACC} 
 & $0.77\!\pm\!0.06$ & $0.76\!\pm\!0.06$ & $0.77\!\pm\!0.06$ 
  & $0.63\!\pm\!0.10$ & $0.62\!\pm\!0.10$ & $0.62\!\pm\!0.10$ 
  & $0.82\!\pm\!0.06$ & $0.82\!\pm\!0.06$ & $0.83\!\pm\!0.06$ 
  
 \\ 
\midrule 

\multirow{2}{*}{\makecell[l]{Random\\Forest}} 
 & \scriptsize{sEMG} 
 & $0.65\!\pm\!0.08$ & $0.68\!\pm\!0.07$ & $\mathbf{0.70\!\pm\!0.07}$ 
  & $0.44\!\pm\!0.08$ & $0.47\!\pm\!0.08$ & $0.48\!\pm\!0.07$ 
  & $0.73\!\pm\!0.06$ & $0.75\!\pm\!0.06$ & $0.77\!\pm\!0.06$ 
  
 \\ 
 & \scriptsize{sEMG + ACC} 
 & $0.86\!\pm\!0.05$ & $0.86\!\pm\!0.05$ & $0.87\!\pm\!0.05$ 
  & $0.73\!\pm\!0.09$ & $0.74\!\pm\!0.09$ & $0.75\!\pm\!0.09$ 
  & $0.90\!\pm\!0.06$ & $0.90\!\pm\!0.06$ & $0.91\!\pm\!0.06$ 
  
 \\ 
\midrule 

\multirow{2}{*}{ShallowCNN} 
 & \scriptsize{sEMG} 
 & $0.61\!\pm\!0.06$ & $0.60\!\pm\!0.06$ & $0.57\!\pm\!0.06$ 
  & $0.40\!\pm\!0.07$ & $0.40\!\pm\!0.07$ & $0.38\!\pm\!0.07$ 
  & $0.68\!\pm\!0.07$ & $0.66\!\pm\!0.07$ & $0.65\!\pm\!0.07$ 
  
 \\ 
 & \scriptsize{sEMG + ACC} 
 & ---  & ---  & ---  
  & ---  & ---  & ---  
  & ---  & ---  & ---  
  
 \\ 
\midrule 

\multirow{2}{*}{ResNet-1D} 
 & \scriptsize{sEMG} 
 & $0.60\!\pm\!0.07$ & $0.61\!\pm\!0.07$ & $0.62\!\pm\!0.07$ 
  & $0.39\!\pm\!0.09$ & $0.41\!\pm\!0.08$ & $0.42\!\pm\!0.09$ 
  & $0.69\!\pm\!0.08$ & $0.69\!\pm\!0.07$ & $0.70\!\pm\!0.07$ 
  
 \\ 
 & \scriptsize{sEMG + ACC} 
 & ---  & ---  & ---  
  & ---  & ---  & ---  
  & ---  & ---  & ---  
  
 \\ 
\midrule 

\multirow{2}{*}{MKCNN} 
 & \scriptsize{sEMG} 
 & $0.66\!\pm\!0.07$ & $0.68\!\pm\!0.07$ & $0.69\!\pm\!0.06$ 
  & $0.45\!\pm\!0.09$ & $0.46\!\pm\!0.08$ & $0.48\!\pm\!0.08$ 
  & $0.74\!\pm\!0.07$ & $0.75\!\pm\!0.07$ & $0.76\!\pm\!0.07$ 
  
 \\ 
 & \scriptsize{sEMG + ACC} 
 & ---  & ---  & ---  
  & ---  & ---  & ---  
  & ---  & ---  & ---  
  
 \\ 
\midrule 

\multirow{2}{*}{NKDFF} 
 & \scriptsize{sEMG} 
 & $0.59\!\pm\!0.11$ & $0.63\!\pm\!0.10$ & $0.66\!\pm\!0.10$ 
  & $0.40\!\pm\!0.09$ & $0.41\!\pm\!0.14$ & $0.44\!\pm\!0.13$ 
  & $0.63\!\pm\!0.11$ & $0.69\!\pm\!0.11$ & $0.65\!\pm\!0.16$ 
  
 \\ 
 & \scriptsize{sEMG + ACC} 
 & $0.86\!\pm\!0.04$ & $0.87\!\pm\!0.04$ & $0.87\!\pm\!0.04$ 
  & $0.72\!\pm\!0.08$ & $0.75\!\pm\!0.09$ & $0.75\!\pm\!0.09$ 
  & $0.89\!\pm\!0.06$ & $0.89\!\pm\!0.07$ & $0.90\!\pm\!0.06$ 
  
 \\ 
\midrule 

\multirow{2}{*}{$\text{EMG-CF}_{\text{1D}}$} 
 & \scriptsize{sEMG} 
 & $0.64\!\pm\!0.08$ & $0.66\!\pm\!0.08$ & $0.68\!\pm\!0.08$ 
  & $0.44\!\pm\!0.09$ & $0.45\!\pm\!0.11$ & $0.49\!\pm\!0.09$ 
  & $0.73\!\pm\!0.07$ & $0.76\!\pm\!0.07$ & $0.76\!\pm\!0.07$ 
  
 \\ 
 & \scriptsize{sEMG + ACC} 
 & $0.89\!\pm\!0.04$ & $0.90\!\pm\!0.04$ & $0.89\!\pm\!0.04$ 
  & $0.78\!\pm\!0.08$ & $0.78\!\pm\!0.08$ & $0.78\!\pm\!0.10$ 
  & $\mathbf{0.92\!\pm\!0.04}$ & $\mathbf{0.92\!\pm\!0.04}$ & $\mathbf{0.92\!\pm\!0.05}$
  
 \\ 
\midrule 

\multirow{2}{*}{$\text{EMG-CF}_{\text{2D}}$} 
 & \scriptsize{sEMG} 
  & $\mathbf{0.68\!\pm\!0.07}$ & $\mathbf{0.69\!\pm\!0.07}$ & $\mathbf{0.70\!\pm\!0.07}$ 
  & $\mathbf{0.46\!\pm\!0.09}$ & $\mathbf{0.48\!\pm\!0.10}$ & $\mathbf{0.50\!\pm\!0.08}$ 
  & $\mathbf{0.76\!\pm\!0.07}$ & $\mathbf{0.77\!\pm\!0.07}$ & $\mathbf{0.78\!\pm\!0.07}$ 
  
 \\ 
 & \scriptsize{sEMG + ACC} 
  & $\mathbf{0.90\!\pm\!0.04}$ & $\mathbf{0.90\!\pm\!0.03}$ & $\mathbf{0.90\!\pm\!0.04}$ 
  & $\mathbf{0.79\!\pm\!0.07}$ & $\mathbf{0.80\!\pm\!0.06}$ & $\mathbf{0.79\!\pm\!0.06}$ 
  & $0.92\!\pm\!0.05$ & $\mathbf{0.92\!\pm\!0.04}$ & $\mathbf{0.92\!\pm\!0.05}$ 
  
 \\ 

\bottomrule
\end{tabular}
\label{tab:cohen}
\end{center}
\captionsetup{hypcap=true}

\end{document}